\documentclass{article}
\PassOptionsToPackage{numbers, compress}{natbib}

\usepackage[final]{neurips_2018}

\usepackage[utf8]{inputenc} 
\usepackage[T1]{fontenc}    
\usepackage{hyperref}       
\usepackage{url}            
\usepackage{booktabs}       
\usepackage{amsfonts}       
\usepackage{nicefrac}       
\usepackage{microtype}      
\usepackage{float}
\usepackage{graphicx}
\usepackage{ownstyles}
\usepackage{verbatim}
\usepackage{multirow}
\usepackage{tabu}
\usepackage{etoolbox}

\graphicspath{{./figs/}}

\newcommand{\capa}{\textbf{(a)}\xspace}
\newcommand{\capb}{\textbf{(b)}\xspace}
\newcommand{\capc}{\textbf{(c)}\xspace}
\newcommand{\capd}{\textbf{(d)}\xspace}
\newcommand{\cape}{\textbf{(e)}\xspace}
\newcommand{\capleft}{\textbf{(left)}\xspace}
\newcommand{\capleftcol}{\textbf{(left column)}\xspace}

\newcommand{\capright}{\textbf{(right)}\xspace}
\newcommand{\caprightcol}{\textbf{(right column)}\xspace}

\newif\ifcomments

\commentsfalse

\ifcomments
\newcommand{\comments}[1]{#1}
\else
\newcommand{\comments}[1]{}
\fi

\usepackage{xr}
\makeatletter
\newcommand*{\addFileDependency}[1]{
  \typeout{(#1)}
  \@addtofilelist{#1}
  \IfFileExists{#1}{}{\typeout{No file #1.}}
}
\makeatother

\newcommand{\titl}{An intriguing failing of convolutional neural networks and the CoordConv solution}
\title{\titl}

    \makeatletter
    \let\@fnsymbol\@arabic
    \makeatother

\author{
  Rosanne Liu$^1$
  \And
  Joel Lehman$^1$
  \And
  Piero Molino$^1$
  \And
  Felipe Petroski Such$^1$
   \And
  Eric Frank$^1$
  \And
  Alex Sergeev$^2$
  \And
  Jason Yosinski$^1$ 
   \AND
   \vspace*{-2.5em}
   ~ \\
   $^1$Uber AI Labs, San Francisco, CA, USA \hspace{1em} $^2$Uber Technologies, Seattle, WA, USA \hspace{1em} \\
   \texttt{\footnotesize \{rosanne,joel.lehman,piero,felipe.such,mysterefrank,asergeev,yosinski\}@uber.com}
}

\begin{document}

\maketitle

\begin{abstract}
  Few ideas have enjoyed as large an impact on deep learning as convolution.
  For any problem involving pixels or spatial representations, common intuition holds that convolutional neural networks may be appropriate.
  In this paper we show a striking counterexample to this intuition via the seemingly trivial \emph{coordinate transform problem}, which simply requires learning a mapping between coordinates in $(x,y)$ Cartesian space and coordinates in one-hot pixel space.
Although convolutional networks would seem appropriate for this task, we show that they fail spectacularly. We demonstrate and carefully analyze the failure first on a toy problem, at which point a simple fix becomes obvious. We call this solution CoordConv, which works by giving convolution access to its own input coordinates through the use of extra coordinate channels. Without sacrificing the computational and parametric efficiency of ordinary convolution, CoordConv allows networks to learn either complete translation invariance or varying degrees of translation dependence, as required by the end task.
CoordConv solves the coordinate transform problem with perfect generalization and 150 times faster with 10--100 times fewer parameters than convolution.
This stark contrast raises the question: to what extent has this inability of convolution persisted insidiously inside other tasks, subtly hampering performance from within? A complete answer to this question will require further investigation, but we show preliminary evidence that swapping convolution for CoordConv can improve models on a diverse set of tasks. Using CoordConv in a GAN produced less mode collapse as the transform between high-level spatial latents and pixels becomes easier to learn. A Faster R-CNN detection model
trained on MNIST detection showed 24\% better IOU when using CoordConv, and
in the Reinforcement Learning (RL) domain agents playing Atari games benefit significantly from the use of CoordConv layers.
\end{abstract}

%

\section{Introduction}
\seclabel{introduction}
\vspace*{-.6em}

Convolutional neural networks (CNNs)
\cite{lecun-1995-convolutional-networks-for-images} have enjoyed
immense success as a key tool for enabling effective deep learning in a
broad array of applications, like modeling natural images
\cite{stack_gan,krizhevsky2012imagenet-classification-with-deep},
images of human faces
\cite{karras-2018-ICLR-progressive-growing-of-gans}, audio
\cite{WaveNet}, and enabling agents to play games in domains with
synthetic imagery like Atari \cite{mnih2013playing-atari-with}.
Although straightforward CNNs excel at many tasks, in many other cases
progress has been accelerated 
through the development of specialized layers that complement the abilities
of CNNs. Detection models like Faster R-CNN \cite{ren-2015-faster-r-cnn:-towards}
make use of layers to compute coordinate transforms and focus attention,
spatial transformer networks \cite{jaderberg:spatial} make use of differentiable cameras
to transform data from the output of one CNN into a form more amenable to processing with another,
and some generative models like DRAW \cite{gregor-2015-draw:-a-recurrent-neural} iteratively
perceive, focus, and refine a canvas rather than using a single pass through a CNN to generate an image.
These models were all created by neural network designers that intuited
some inability or misguided inductive bias of standard CNNs and then
devised a workaround.

In this work, we expose and analyze a generic inability of CNNs
to transform spatial representations between two different types: from a dense Cartesian
representation to a sparse, pixel-based representation or in the opposite direction.
Though such transformations would seem simple for networks to learn, it turns out to be more difficult than expected, at least when models are comprised of the commonly used stacks of convolutional layers. While straightforward stacks of convolutional layers excel at tasks like image classification, they are not quite the right model for coordinate transform.

The main contributions of this paper are as follows:

\figp[t]{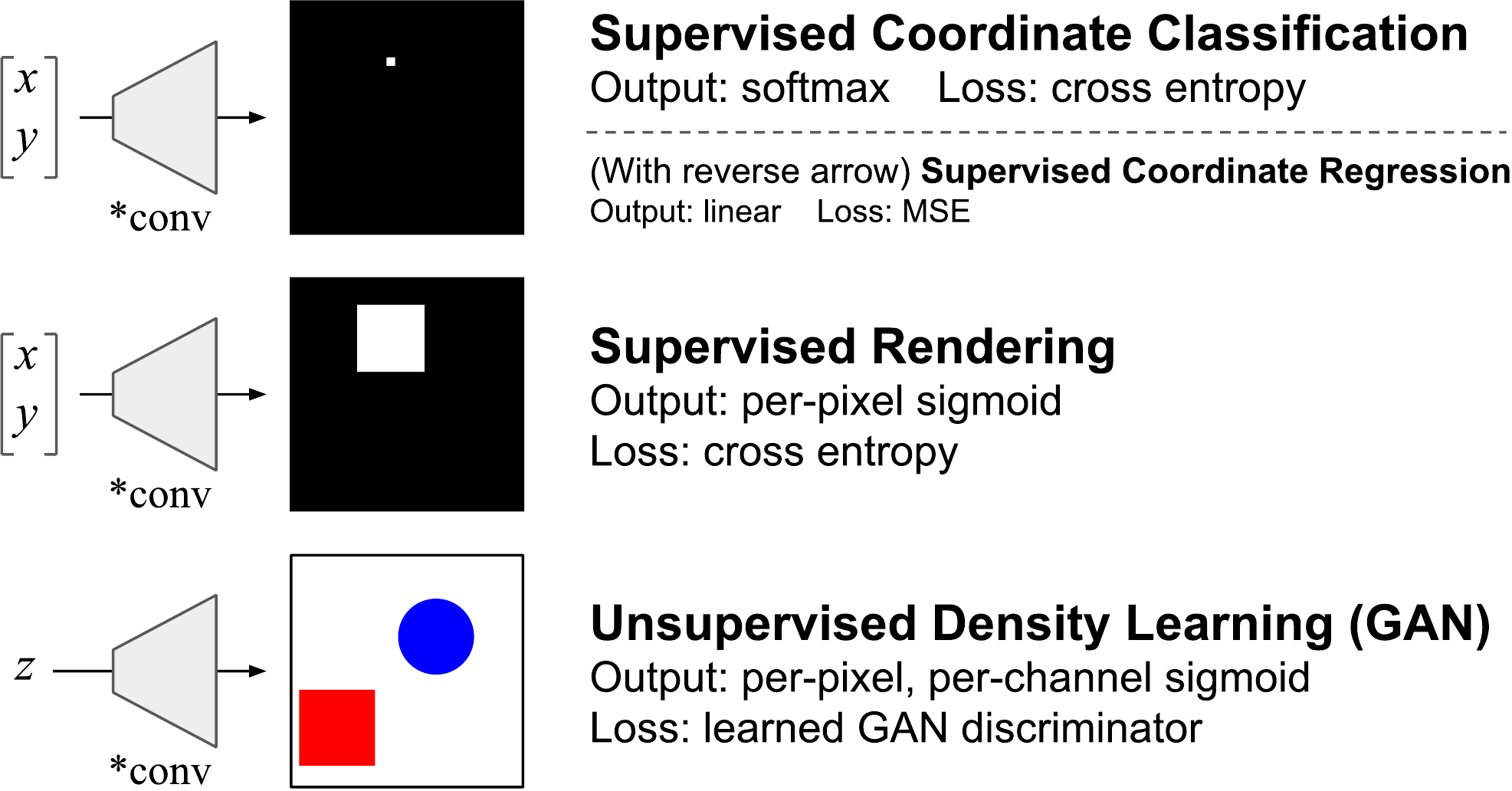}{.6}{
  Toy tasks considered in this paper. The \emph{*conv} block represents a network comprised of one or more convolution, deconvolution (convolution transpose), or CoordConv layers.
  Experiments compare networks with no CoordConv layers to those with one or more.\vspace*{-1em}}

\begin{enumerate}
\item We define a simple toy dataset, \emph{Not-so-Clevr}, which consists of squares randomly positioned on a canvas  (\secref{dataset}).

\item We define the \emph{CoordConv} operation, which allows convolutional filters to know where they are in Cartesian space by adding extra, hard-coded input channels that contain coordinates of the data seen by the convolutional filter. The operation may be implemented via a couple extra lines of Tensorflow (\secref{coordconv}).

\item Throughout the rest of the paper, we examine the coordinate transform problem starting with the simplest scenario and ending with the most complex.
Although results on toy problems should generally be taken with a degree of skepticism, starting small allows us to pinpoint the issue, exploring and understanding it in detail.
Later sections then show that the phenomenon observed in the toy domain indeed appears in more real-world settings.

\vspace{.2em} We begin by showing that coordinate transforms are surprisingly difficult even when the problem is \emph{small and supervised}. In the \emph{Supervised Coordinate Classification} task,
given a pixel's $(x,y)$ coordinates as input, we train a CNN to highlight it as output. The \emph{Supervised Coordinate Regression} task entails the inverse:
given an input image containing a single white pixel, output its coordinates. We show that both problems are harder than expected using convolutional layers but become trivial by using a CoordConv layer (\secref{supervised_classification}).

\item The
  \emph{Supervised Rendering} task adds complexity to the above by requiring a network
  to paint a full image from the Not-so-Clevr dataset given the $(x,y)$ coordinates of the center of a square in the image.
  The task is still fully supervised, but as before, the task is difficult to learn for convolution and trivial for CoordConv (\secref{supervised_rendering}).

\item
  We show that replacing convolutional layers with CoordConv improves performance in a variety of tasks.
  On two-object Sort-of-Clevr \cite{sortofclevr} images, Generative Adversarial Networks (GANs) and Variational Autoencoders (VAEs) using CoordConv exhibit
  less mode collapse, perhaps because ease of learning coordinate transforms translates to ease of using latents to span a 2D Cartesian space. Larger GANs on bedroom scenes with CoordConv offer geometric translation that was never observed in regular GAN.
  Adding CoordConv to a Faster R-CNN produces much better object boxes and scores.
  Finally, agents learning to play Atari games obtain significantly higher scores on some but not all games, and they never do significantly worse
  (\secref{other_results}).

\item
To enable other researchers to reproduce experiments in this paper, 
and benefit from using CoordConv as a simple drop-in replacement of the convolution layer in their models,
we release our code at \url{https://github.com/uber-research/coordconv}.

\end{enumerate}


With reference to the above numbered contributions, the reader may be interested to know that the course of this research originally progressed in the
$5 \rightarrow 2$ direction as we debugged why progressively simpler problems continued to elude straightforward modeling. But for ease of presentation, we give results in the $2 \rightarrow 5$ direction.
A progression of the toy problems considered is shown in \figref{tasks}.

\section{Not-so-Clevr dataset}
\seclabel{dataset}
\vspace*{-.6em}

We define the Not-so-Clevr dataset and make use of it for the first experiments in this paper. The dataset is a single-object, grayscale version of Sort-of-CLEVR \cite{sortofclevr}, which itself is a simpler version of the Clevr dataset of rendered 3D shapes \cite{clevr}.
Note that the series of Clevr datasets have been typically used for studies regarding relations and visual question answering, but we here use them for supervised learning and generative models.
Not-so-Clevr consists of $9\times9$ squares placed on a $64\times64$ canvas. Square positions are restricted such that the entire square lies within the $64\times64$ grid, so that square centers fall within a slightly smaller possible area of $56\times56$. Enumerating these possible center positions results in a dataset with a total of 3,136 examples.
For each example square $i$, the dataset contains three fields:

\vspace*{-.6em}
\begin{itemize}
\item $C_i \in \mathbb{R}^2$, its center location in $(x,y)$ Cartesian coordinates, \vspace*{-.2em}
\item $P_i \in \mathbb{R}^{64\times64}$, a one-hot representation of its center pixel, and \vspace*{-.2em}
\item $I_i \in \mathbb{R}^{64\times64}$, the resulting $64\times64$ image of the square painted on the canvas.
\end{itemize}
\vspace*{-.6em}

We define two train/test splits of these 3,136 examples: \emph{uniform}, where all possible center locations are randomly split 80/20 into train vs. test sets, and \emph{quadrant}, where three of four quadrants are in the train set and the fourth quadrant in the test set. Examples from the dataset and both splits are depicted in \figref{dataset}.
To emphasize the simplicity of the data, we note that this dataset may be generated in only a line or two of Python using a single convolutional layer with filter size $9\times9$ to paint the squares from a one-hot representation.\footnote{For example, ignoring import lines and train/test splits: \\{\scriptsize \code{onehots = np.pad(np.eye(3136).reshape((3136, 56, 56, 1)), ((0,0), (4,4), (4,4), (0,0)), "constant"); \\images = tf.nn.conv2d(onehots, np.ones((9, 9, 1, 1)), [1]*4, "SAME")}}}

\figp[t]{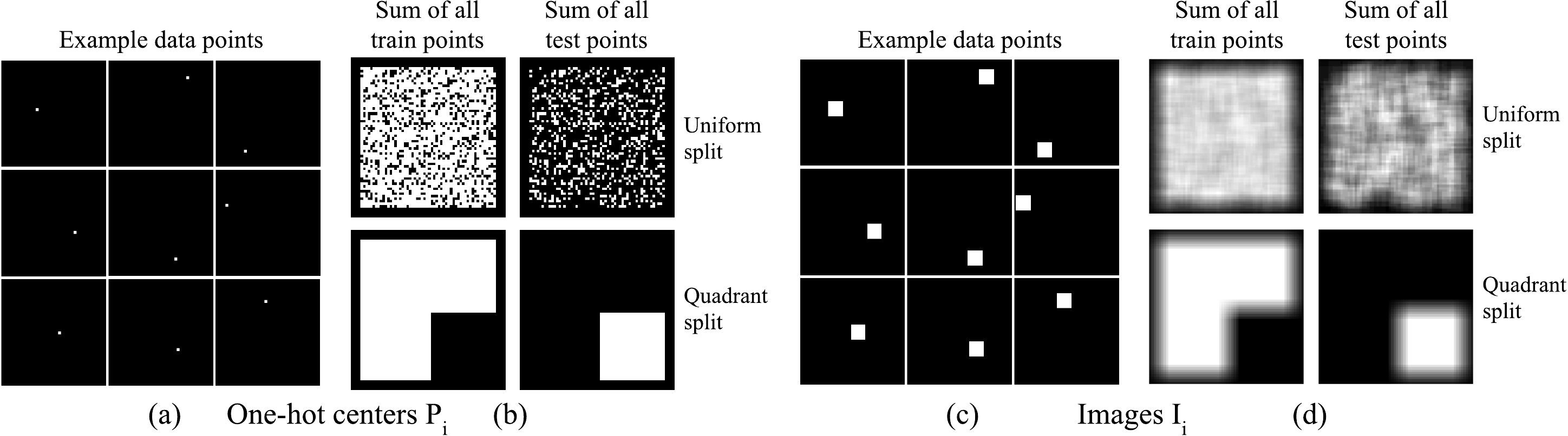}{.9}{
  The Not-so-Clevr dataset.
  \capa Example one-hot center images $P_i$ from the dataset.
  \capb The pixelwise sum of the entire train and test splits for uniform vs. quadrant splits.
  \capc and \capd Analagous depictions of the canvas images $I_i$ from the dataset.
  Best viewed electronically with zoom.\vspace*{-1em}
}


\section{The CoordConv layer}
\seclabel{coordconv}
\vspace*{-.6em}

The proposed CoordConv layer is a simple extension to the standard convolutional layer. We assume for the rest of the paper the case of two spatial dimensions, though operators in other dimensions follow trivially.
Convolutional layers are used in a myriad of applications because they often work well, perhaps due to some combination of three factors: they have relatively few learned parameters, they are fast to compute on modern GPUs, and they learn a function that is translation invariant (a translated input produces a translated output).

The CoordConv layer keeps the first two of these properties---few parameters and efficient computation---but allows the network to learn to keep or to discard the third---translation invariance---as is needed for the task being learned.
It may appear that doing away with translation invariance will hamper networks' abilities to learn generalizable functions. However, as we will see in later sections, allocating a small amount of network capacity to model non-translation invariant aspects of a problem can enable far more trainable models that also generalize far better.

The CoordConv layer can be implemented as a simple extension of standard convolution in which extra channels are instantiated and filled with (constant, untrained) coordinate information, after which they are  concatenated channel-wise to the input representation and a standard convolutional layer is applied.
\figref{coordconv} depicts the operation where two coordinates, $i$ and $j$, are added.
Concretely, the $i$ coordinate channel is an $h\times w$ rank-1 matrix with its first row filled with 0's, its second row with 1's, its third with 2's, etc. The $j$ coordinate channel is similar, but with columns filled in with constant values instead of rows. In all experiments, we apply a final linear scaling of both $i$ and $j$ coordinate values to make them fall in the range $[-1,1]$.
For convolution over two dimensions, two $(i,j)$ coordinates are sufficient to completely specify an input pixel, but if desired, further channels can be added as well to bias models toward learning particular solutions.
In some of the experiments that follow, we have also used a third channel for an $r$ coordinate, where $r = \sqrt{(i-h/2)^2+(j-w/2)^2}$. The full implementation of the CoordConv layer is provided in \secref{si_layer}.
Let's consider next a few properties of this operation.

\paragraph{Number of parameters.} \vspace*{-.6em}
Ignoring bias parameters (which are not changed), a standard convolutional layer with square kernel size $k$ and with $c$ input channels and $c'$ output channels will contain $cc'k^2$ weights, whereas the corresponding CoordConv layer will contain $(c+d)c'k^2$ weights, where $d$ is the number of coordinate dimensions used (e.g. 2 or 3). The relative increase in parameters is small to moderate, depending on the original number of input channels.
\footnote{A CoordConv layer implemented via the channel concatenation discussed entails an increase of $dc'k^2$ weights. However, if $k>1$, not all $k^2$ connections from coordinates to each output unit are necessary, as spatially neighboring coordinates do not provide new information. Thus, if one cares acutely about minimizing the number of parameters and operations, a $k\times k$ conv may be applied to the input data and a $1\times1$ conv to the coordinates, then the results added. In this paper we have used the simpler, if marginally inefficient, channel concatenation version that applies a single convolution to both input data and coordinates. However, almost all experiments use $1\times1$ filters with CoordConv.}

\figp[t]{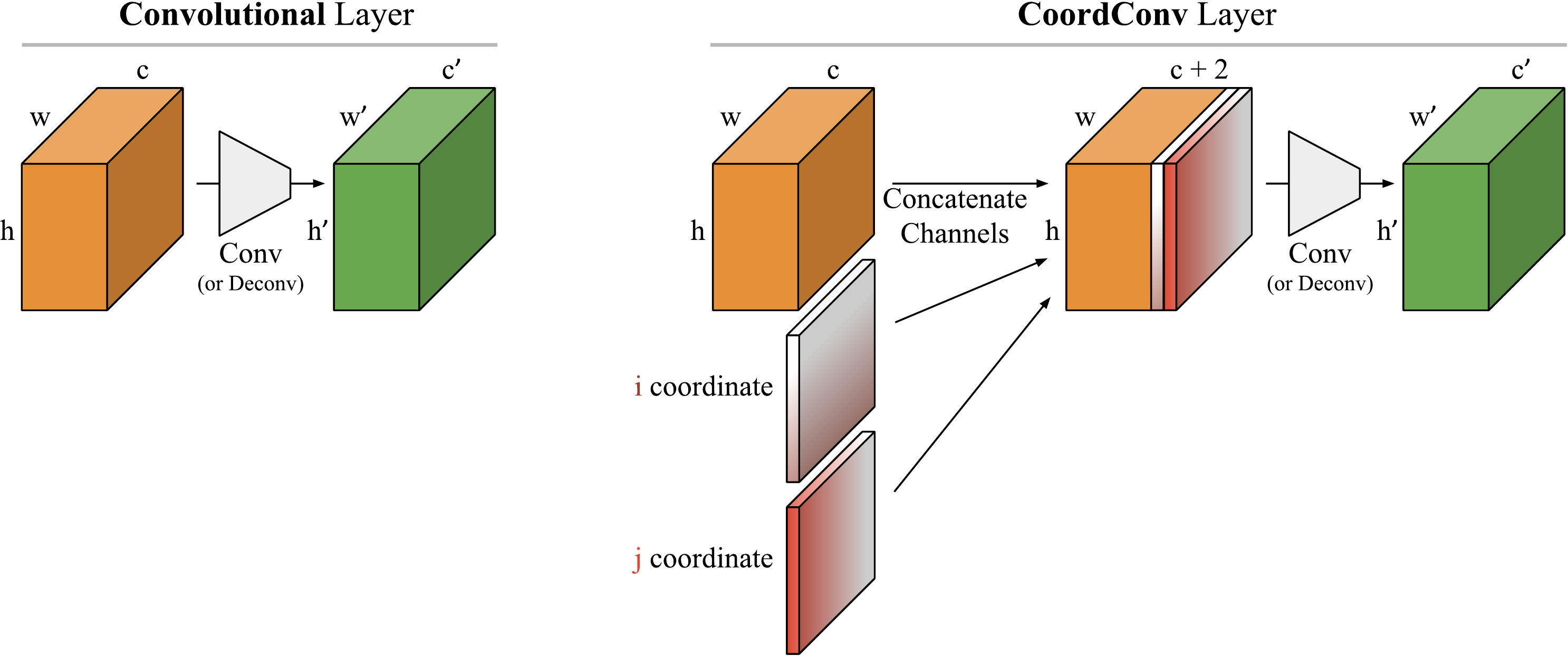}{.8}{Comparison of 2D convolutional and CoordConv layers.
\capleft A standard convolutional layer maps from a representation block with shape $h\times w \times c$ to a new representation of shape $h'\times w' \times c'$.
\capright A CoordConv layer has the same functional signature, but accomplishes the mapping by first concatenating extra channels to the incoming representation. These channels contain hard-coded coordinates, the most basic version of which is one channel for the $i$ coordinate and one for the $j$ coordinate, as shown above. Other derived coordinates may be input as well, like the radius coordinate used in ImageNet experiments (\secref{other_results}).
\vspace*{-1em}
}

\paragraph{Translation invariance.} \vspace*{-.6em}
CoordConv with weights connected to input coordinates set by initialization or learning to zero will be translation invariant and thus mathematically equivalent to ordinary convolution.
 If weights are nonzero, the function will contain some degree of translation dependence, the precise form of which will ideally depend on the task being solved.
 Similar to locally connected layers with unshared weights, CoordConv allows learned translation dependence, but by contrast it requires far fewer parameters: $(c+d)c'k^2$ vs. $hwcc'k^2$ for spatial input size $h\times w$.
Note that all CoordConv weights, even those to coordinates, are shared across all positions, so 
translation dependence comes only from the specification of coordinates; one consequence is that, as with ordinary convolution but unlike locally connected layers, the operation can be expanded outside the original spatial domain if the appropriate coordinates are extrapolated.

\paragraph{Relations to other work.} \vspace*{-0.6em}
CoordConv layers are related to many other bodies of work.
Compositional Pattern Producing Networks (CPPNs)
\cite{stanley-2007-GPEM-compositional-pattern-producing}
implement functions from coordinates in arbitrarily many dimensions to one or more output values. For example, with two input dimensions and $N$ output values, this can be thought of as painting $N$ separate grayscale pictures.
CoordConv can then be thought of as a conditional CPPN where output values depend not only on coordinates but also on incoming data.
In one CPPN-derived work \cite{hoover-2009-ConnSci-exploiting-functional-relationships}, researchers did train networks that take as input both coordinates and incoming data for their use case of synthesizing a drum track that could derive both from a time coordinate and from other instruments (input data) and trained using interactive evolution. With respect to that work, we may see CoordConv as a simpler, single-layer mechanism that fits well within the paradigm of training large networks with gradient descent on GPUs.
In a similar vein, research on convolutional sequence to sequence learning \cite{gehring-2017-arXiv-convolutional-sequence-to-sequence} has used fixed and learned position embeddings at the input layer; in that work, positions were represented via an overcomplete basis that is added to the incoming data rather than being compactly represented and input as separate channels. In some cases using overcomplete sine and cosine bases or learned encodings for locations has seemed to work well \cite{vaswani-2017-NIPS-attention-is-all-you-need, parmar2018image}.
Connections can also be made to mechanisms of spatial attention \cite{jaderberg:spatial} and to generative models that separately learn what and where to draw \cite{gregor-2015-draw:-a-recurrent-neural, whatwhere_gan}. While such works might appear to provide alternative solutions to the problem explored in this paper, in reality, similar coordinate transforms are often embedded within such models (e.g.\ a spatial transformer network contains a localization network that regresses from an image to a coordinate-based representation \cite{jaderberg:spatial}) and might also benefit from CoordConv layers. 

Moreover, several previous works have found it necessary or useful to inject geometric information to networks, for example, in prior networks to enhance spatial smoothness \cite{ulyanov2017deep}, in segmentation networks \cite{Brust15:CPN,lyu2018road}, and in robotics control through a spatial softmax layer and an expected coordinate layer that map scenes to object locations \cite{levine2016end, finn2016deep}. However, in those works it is often seen as a minor detail in a larger architecture which is tuned to a specific task and experimental project, 
and discussions of this necessity are scarce. 
In contrast, our research 
(a) examines this necessity in depth as its central thrust,
(b) reduces the difficulty to its minimal form (coordinate transform), leading to a simple single explanation that unifies previously disconnected observations, and
(c) presents one solution used in various forms by others as a unified layer, easily included anywhere in any convolutional net.
Indeed, the wide range of prior works provide strong evidence of the generality of the core coordinate transform problem 
across domains, suggesting the significant value of a work that systematically explores its impact and
collects together these disparate previous references.

Finally, we note that challenges in learning coordinate transformations are not unknown in machine learning, as learning a Cartesian-to-polar coordinate transform forms the basis of the classic two-spirals classification problem \cite{fahlman:cascade}.




\figp[t]{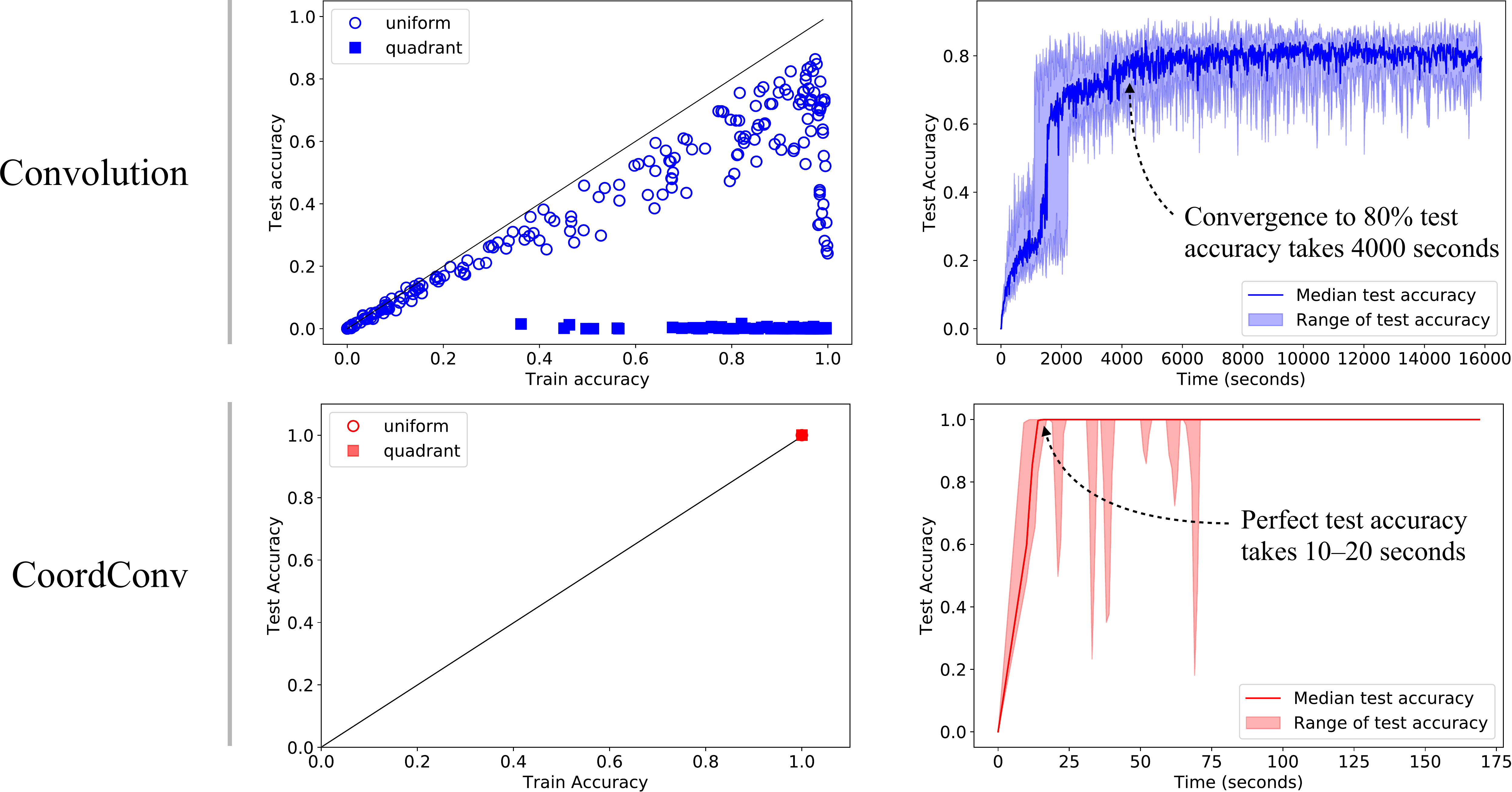}{.8}{
Performance of convolution and CoordConv on Supervised Coordinate Classification.
\capleftcol Final test vs. train accuracy. On the easier uniform split, convolution never attains perfect test accuracy, though the largest models memorize the training set. On the quadrant split, generalization is almost zero.
CoordConv attains perfect train and test accuracy on both splits.
One of the main results of this paper is that the translation invariance in ordinary convolution does not lead to coordinate transform generalization even to neighboring pixels! 
\caprightcol Test accuracy vs. training time of the best uniform-split models from the left plot (any reaching final test accuracy $\geq 0.8$). The convolution models never achieve more than about 86\% accuracy, and training is slow: the fastest learning models still take over an hour to converge. 
CoordConv models learn several hundred times faster, attaining perfect accuracy in seconds.\vspace*{-1em}
%
%
\vspace*{-1em}
}

\vspace*{-.5em}
\section{Supervised Coordinate tasks}
\seclabel{supervised_classification}
\vspace*{-.6em}

\subsection{Supervised Coordinate Classification}
\seclabel{supervised_classification1}
\vspace*{-.6em}

The first and simplest task we consider is Supervised Coordinate Classification.
Illustrated at the top of \figref{tasks}, given an $(x,y)$ coordinate as input, a network must learn to paint the correct output pixel. This is simply a multi-class classification problem where each pixel is a class.
Why should we study such a toy problem? If we expect to train generative models that can transform high level latents like horizontal and vertical position into pixel data, solving this toy task would seem a simple prerequisite. We later verify that performance on this task does in fact predict performance on larger problems.

In \figref{supcoord_classification_crop} we depict training vs. test accuracy on the task for both uniform and quadrant train/test splits.
For convolutional models\footnote{For classification, convolutions and CoordConvs are actually deconvolutional on certain layers when resolutions must be expanded, but we refer to the models as conv or CoordConv for simplicity.}(6 layers of deconvolution with stride 2, see  \secref{si_model_arch} in the Supplementary Information for architecture details) on uniform splits, we find models that generalize somewhat, but 100\% test accuracy is never achieved, with the best model achieving only 86\% test accuracy. This is surprising: because of the way the uniform train/test splits were created, all test pixels are close to multiple train pixels.
Thus, we reach a first striking conclusion: \emph{learning a smooth function from $(x,y)$ to one-hot pixel is difficult for convolutional networks, even when trained with supervision, and even when supervision is provided on all sides.}
Further, training a convolutional model to 86\% accuracy takes over an hour and requires about 200k parameters (see \secref{si_supervised_coordinate} in the Supplementary Information for details on training).
On the quadrant split, convolutional models are unable to generalize at all. \figref{supcoord_sumresults_both_crop} shows sums over training set and test set predictions, showing visually both the memorization of the convolutional model and its lack of generalization.

In striking contrast, CoordConv models attain perfect performance on both data splits and do so with only 7.5k parameters and in only 10--20 seconds.
The parsimony of parameters further confirms they are simply more appropriate models for the task of coordinate transform \cite{rissanen-1978-automatica-modeling-by-shortest-data,hinton-1993-COLT-keeping-the-neural-networks,li-2018-ICLR-measuring-the-intrinsic-dimension}.



\figp[t]{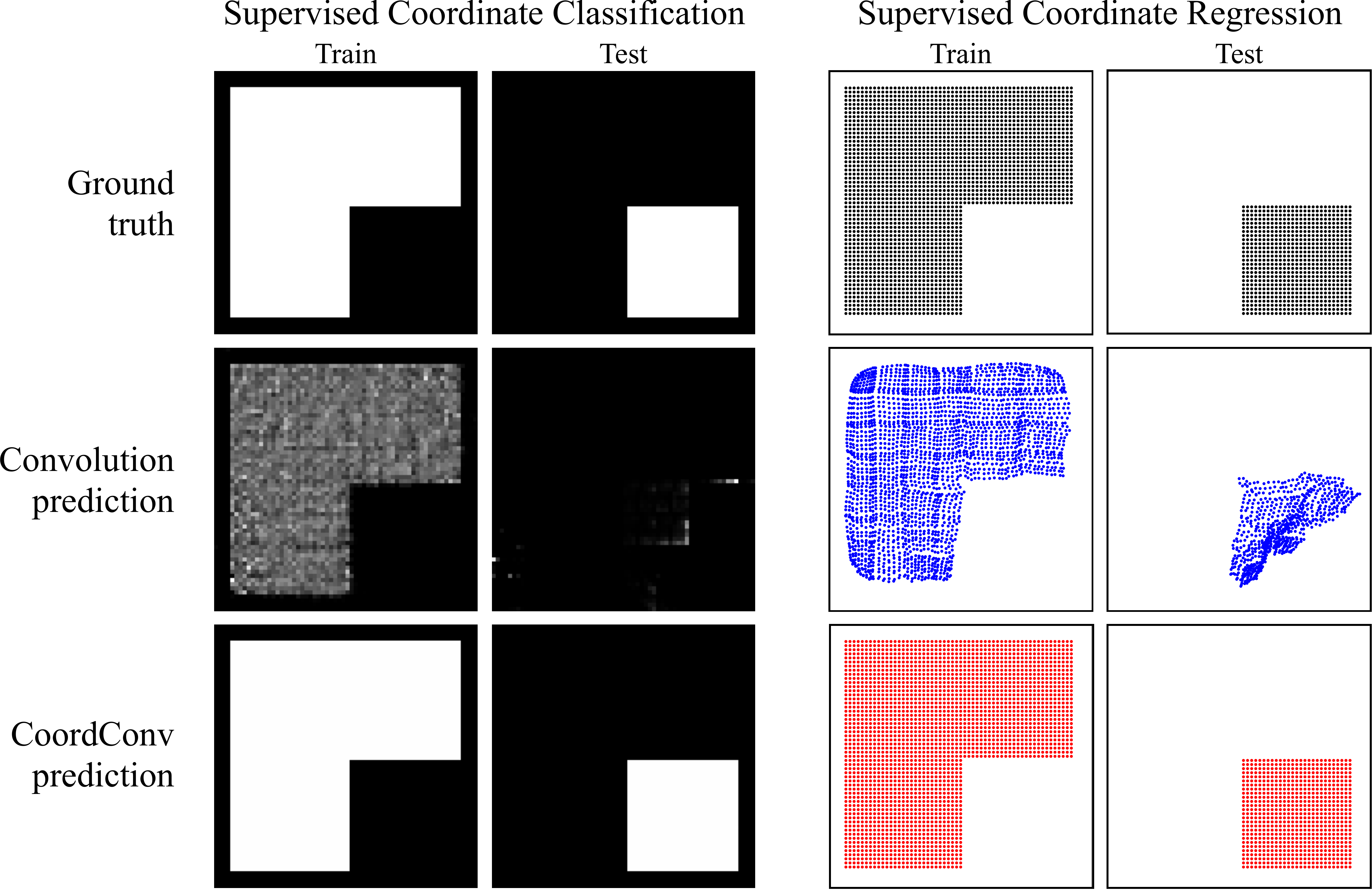}{.80}{
  Comparison of convolutional and CoordConv models on the Supervised Coordinate Classification and Regression tasks, on a quadrant split.
  \capleftcol Results on the seemingly simple classification task where the network must highlight one pixel given its $(x,y)$ coordinates as input.
  Images depict ground truth or predicted probabilities summed across the entire train or test set and then normalized to make use of the entire black to white image range. Thus, e.g., the top-left image shows the sum of all train set examples. The conv predictions on the train set cover it well, although the amount of noise in predictions hints at the difficulty with which this model eventually attained 99.6\% train accuracy by memorization. The conv predictions on the test set are almost entirely incorrect, with two pixel locations capturing the bulk of the probability for all locations in the test set. By contrast, the CoordConv model attains 100\% accuracy on both the train and test sets. Models used: conv--6 layers of deconv with strides 2; CoordConv--5 layers of 1$\times$1 conv, first layer is CoordConv. Details in \secref{si_supervised_coordinate}.
  \caprightcol The regression task poses the inverse problem: predict real-valued $(x,y)$ coordinates from a one-hot pixel input. As before, the conv model memorizes poorly and largely fails to generalize, while the CoordConv model fits train and test set perfectly. Thus we observe the coordinate transform problem to be difficult in both directions.
  Models used: conv--9-layer fully-convolution with global pooling; CoordConv--5 layers of conv with global pooling, first layer is CoordConv. Details in \secref{si_supervised_regress}.
  \vspace*{-1.5em}
}






\vspace*{-.6em}
\subsection{Supervised Coordinate Regression}
\seclabel{supervised_regression}
\vspace*{-.6em}

Because of the surprising difficulty of learning to transform coordinates from Cartesian to a pixel-based, we examine whether the inverse transformation from pixel-based to Cartesian is equally difficult.
This is the type of transform that could be employed by a VAE encoder or GAN discriminator to transform pixel information into higher level latents encoding locations of objects in a scene.

We experimented with various convolutional network structures, and found a 4-layer convolutional network with fully connected layers (85k parameters, see \secref{si_supervised_regress} for details) can fit the uniform training split
and generalize well (less than half a pixel error on average), but that same architecture completely fails on the quadrant split. 
A smaller fully-convolutional architecture (12k parameters, see \secref{si_supervised_regress}) can be tuned to achieve limited
generalization on the quadrant split (around five pixels error on average) as shown in \figref{supcoord_sumresults_both_crop} (right column), but it performs poorly on the uniform split.

A number of factors may have led to the observed variation of performance, including the use of max-pooling, batch normalization, and fully-connected layers.
We have not fully and separately measured how much each factor contributes to poor performance on these tasks;
rather we report only that our efforts to find a workable architecture across both splits did not yield any winners.
In contrast, a 900 parameter CoordConv model, where a single CoordConv layer is followed by several layers of standard convolution,
trains quickly and generalizes in both the uniform and quadrant splits. See \secref{si_supervised_regress} in Supplementary Information for more details. These results suggest that the inverse transformation requires similar considerations to solve as the Cartesian-to-pixel transformation.






\vspace*{-.6em}
\subsection{Supervised Rendering}
\seclabel{supervised_rendering}
\vspace*{-.6em}

Moving beyond the domain of single pixel coordinate transforms, we compare performance of convolutional vs. CoordConv networks on the Supervised Rendering task, which requires a network to produce a $64\times64$ image with a square painted centered at the given $(x,y)$ location. 
As shown in \figref{suppaint_deconv_coordconv_trainval}, we observe the same stark contrast between convolution and CoordConv.
Architectures used for both models can be seen in \secref{si_model_arch} in the Supplementary Information, along with further plots, details of training, and hyperparameter sweeps given in \secref{si_supervised_rendering}.

\figgp[t]{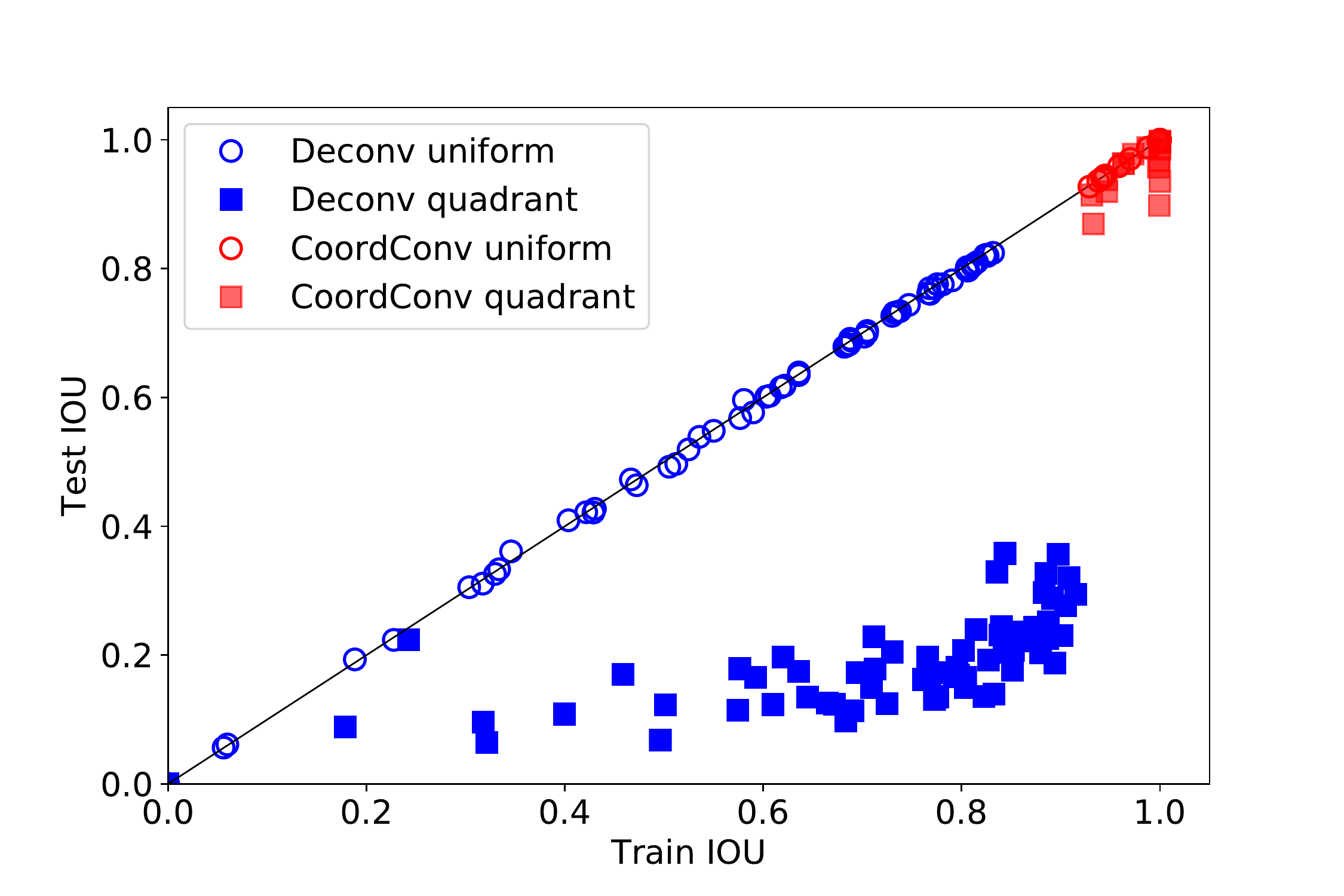}{.4}{suppaint_deconv_coordconv_time_together}{.59}{
  Results on the Supervised Rendering task.
  As with the Supervised Coordinate Classification and Regression tasks, we see the same vast
  separation in training time and generalization between convolution models and CoordConv models.
  \capleft
  Test intersection over union (IOU) vs Train IOU. We show all attempted models on the uniform and quadrant splits, including some CoordConv models whose hyperparameter selections led to worse than perfect performance.
  \capright
  Test IOU vs. training time
  of the best uniform-split models from the left plot (any reaching final test IOU $\geq 0.8$).
  Convolution models never achieve more than about IOU 0.83, and training is slow: the fastest learning models still take over two hours to converge vs. about a minute for CoordConv models.
  %
  \vspace*{-1em}
}



\section{Applicability to Image Classification, Object Detection, Generative Modeling, and Reinforcement Learning}
\seclabel{other_results}

Given the starkly contrasting results above, it is natural to ask how much the demonstrated inability of convolution
at coordinate transforms infects other tasks.
Does the coordinate transform hurdle persist insidiously inside other tasks, subtly hampering performance from within?
Or do networks skirt the issue by learning around it, perhaps by representing space differently, e.g. via non-Cartesian representations like grid cells 
\cite{banino-2018-Nature-vector-based-navigation-using,franzius-2007-PLoSCB-slowness-and-sparseness-lead,cueva-2018-ICLR-emergence-of-grid-like-representations}?
A complete answer to this question is beyond the scope of this paper, but encouraging preliminary evidence shows that
swapping Conv for CoordConv can improve a diverse set of models --- including ResNet-50, Faster R-CNN, GANs, VAEs, and RL models.


\paragraph{ImageNet Classification}
As might be expected for tasks requiring straightforward translation \textit{invariance}, CoordConv does not help significantly when tested with image classification. Adding a single extra 1$\times$1 CoordConv layer with 8 output channels improves ResNet-50 \cite{he-2015-arXiv-deep-residual-learning} Top-5 accuracy by a meager 0.04\% averaged over five runs for each treatment; however, this difference is not statistically significant. It is at least reassuring that CoordConv doesn't hurt the performance since it can always learn to ignore coordinates.
This result was obtained using distributed training on 100 GPUs with Horovod~\cite{sergeev-2018-arXiv-horovod:-fast-and-easy}; see \secref{si_imagenet} in Supplementary Information for more details.

\vspace*{-.5em}

\paragraph{Object Detection} 
In object detection, models look at pixel space and output bounding boxes in Cartesian space. This creates a natural coordinate transform problem which makes CoordConv seemingly a natural fit. On a simple problem of detecting MNIST digits scattered on a canvas, we found the test intersection-over-union (IOU) of a Faster R-CNN network improved by 24\% when using CoordConv. See \secref{si_objdet} in Supplementary Information for details.

\vspace*{-.5em}

\paragraph{Generative Modeling}

Well-trained generative models can generate visually compelling images \cite{nguyen-2016-arXiv-plug-amp-play-generative,karras-2018-ICLR-progressive-growing-of-gans,stack_gan}, but careful inspection can reveal mode collapse: images are of an attractive quality, but sample diversity is far less than diversity present in the dataset. Mode collapse can occur in many dimensions, including those having to do with content, style, or position of components of a scene. We hypothesize that mode collapse of position may be due to the difficulty of learning straightforward transforms from a high-level latent space containing coordinate information to pixel space and that using CoordConv could help. First we investigate a simple task of generating colored shapes with, in particular, all possible geometric locations, using both GANs and VAEs. Then we scale up the problem to Large-scale Scene Understanding (LSUN) \cite{lsun} bedroom scenes with DCGAN~\cite{dcgan}, through distributed training using Horovod~\cite{sergeev-2018-arXiv-horovod:-fast-and-easy}.

Using GANs to generate simple colored objects,  \figref{gan_samples}a-d show sampled images and model collapse analyses. We observe that a convolutional GAN exhibits collapse of a two-dimensional distribution to a one-dimensional manifold. The corresponding CoordConv GAN model generates objects that better cover the 2D Cartesian space while using 7\% of the parameters of the conv GAN. Details of the dataset and training can be seen in \secref{si_gan_clevr} in the Supplementary Information. A similar story with VAEs is discussed in \secref{si_vae}. 

With LSUN, samples are shown in \figref{gan_samples}e, and more in \secref{si_gan_lsun} in the Supplementary Information. We observe (1) qualitatively comparable samples when drawing randomly from each model, and (2) geometric translating behavior during latent space interpolation. 

Latent space interpolation\footnote{https://www.youtube.com/watch?v=YefMbLqS7Jg} demonstrates that in generating colored objects,  motions through latent space generate coordinated object motion. In LSUN, while with convolution we see frozen objects fading in and out, with CoordConv, we instead see smooth geometric transformations including translation and deformation.

\figp[t]{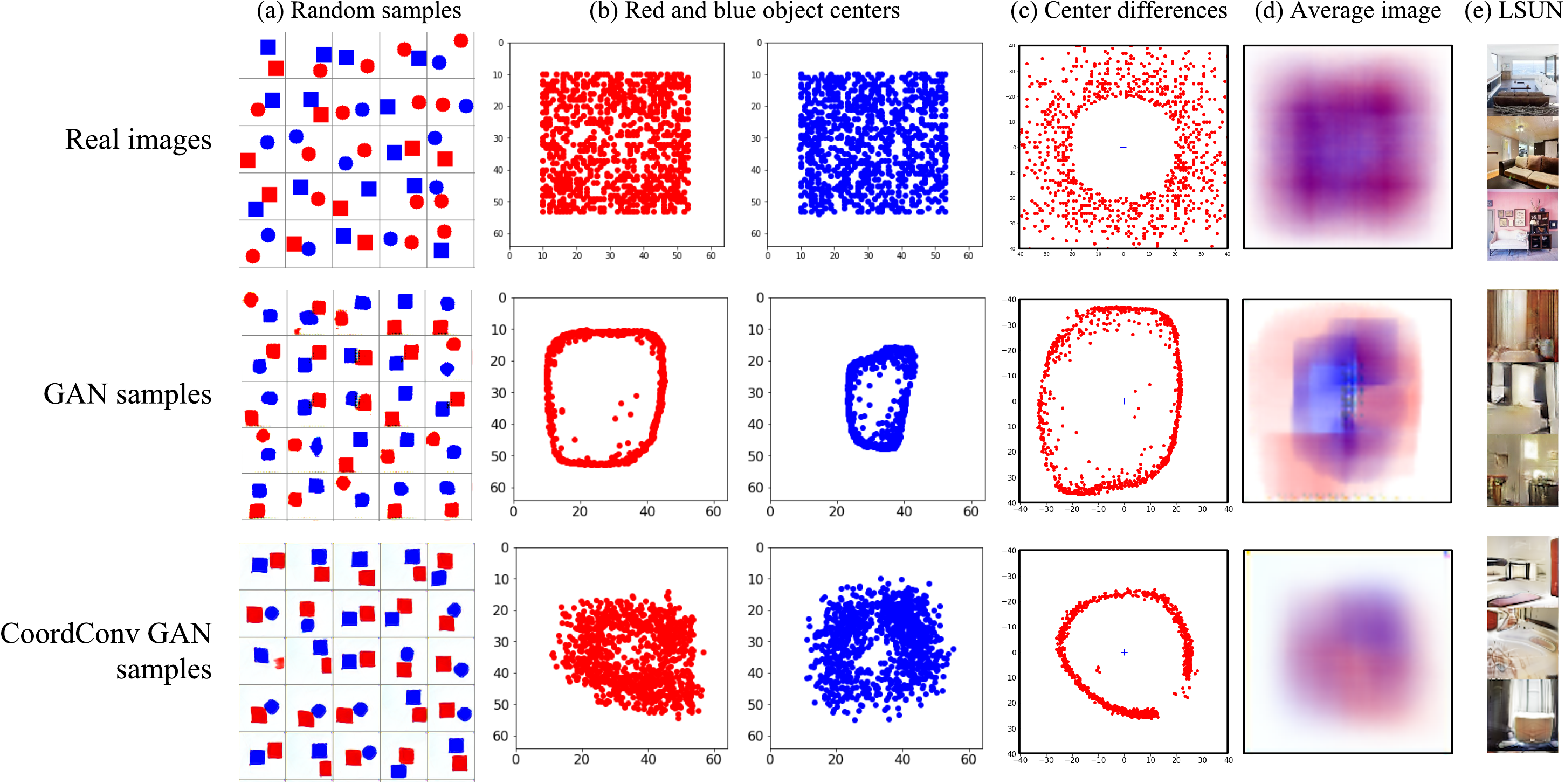}{.9} {Real images and generated images by GAN and CoordConv GAN. Both models learn the basic concepts similarly well: two objects per image, one red and one blue, their size is fixed, and their positions can be random \capa. 
However, plotting the spread of object centers over 1000 samples, we see that CoordConv GAN samples cover the space significantly better (average entropy: \textit{Data} red 4.0, blue 4.0, diff 3.5; \textit{GAN} red 3.13, blue 2.69, diff 2.81; \textit{CoordConv GAN} red 3.30, blue 2.93, diff 2.62), while GAN samples exhibit mode collapse on where objects can be \capb. In terms of relative locations between the two objects, both model exhibit a certain level of model collapse, CoordConv is worse \capc. The averaged image of CoordConv GAN is smoother and closer to that of data \capd. With LSUN, sampled images are shown \cape. All models used in generation are the best out of many runs.}



\vspace*{-.5em}
\paragraph{Reinforcement Learning}
Adding a CoordConv layer to an actor network within A2C \cite{mnih:a3c} produces significant improvements on some games, but not all, as shown in \figref{rl_results_9games}. We also tried adding CoordConv to our own implementation of Distributed Prioritized Experience Replay (Ape-X) \cite{horgan:apex}, but we did not notice any immediate difference. Details of training are included in \secref{si_rl}.

\figp[t]{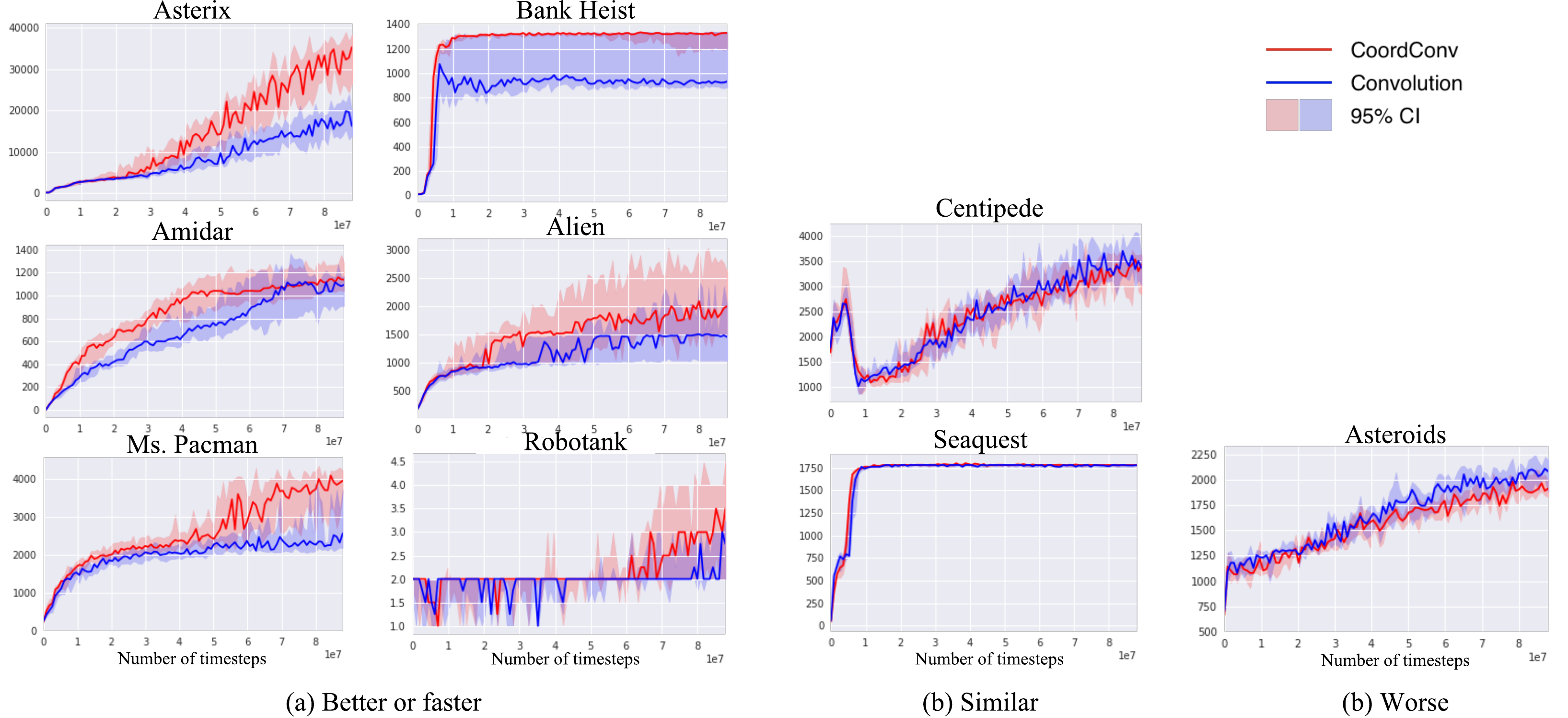}{.9}{Results using A2C to train on Atari games. Out of 9 games, (a) in 6 CoordConv improves over convolution, (b) in 2 performs similarly, and (c) on 1 it is slightly worse.\vspace*{-1em}} 

\vspace*{-.5em}
\section{Conclusions and Future Work}
\seclabel{conclusion}
\vspace*{-.6em}

We have shown the curious inability of CNNs to model the coordinate transform task, shown a simple fix in the form
of the CoordConv layer, and given results that suggest including these layers can boost performance
in a wide range of applications. Future work will further evaluate the benefits of CoordConv in large-scale datasets, 
exploring its ability against perturbations of translation, 
its impact in relational reasoning \cite{sortofclevr}, language tasks, video prediction, with spatial transformer networks \cite{jaderberg:spatial}, and with cutting-edge generative models \cite{gregor-2015-draw:-a-recurrent-neural}.

\clearpage

\section*{Acknowledgements}
The authors gratefully acknowledge Zoubin Ghahramani, Peter Dayan, and Ken Stanley for insightful discussions. 
We are also grateful to the entire Opus team and Machine Learning Platform team inside Uber for providing our computing platform and for technical support.

\bibliography{jby_refs,other_refs}
\bibliographystyle{plain}

\clearpage

\renewcommand{\thesection}{S\arabic{section}}
\renewcommand{\thesubsection}{\thesection.\arabic{subsection}}

\newcommand{\beginsupplementary}{%
		\setcounter{table}{0}
	\renewcommand{\thetable}{S\arabic{table}}%
		\setcounter{figure}{0}
	\renewcommand{\thefigure}{S\arabic{figure}}%
		\setcounter{section}{0}
}
     
\beginsupplementary

\noindent\makebox[\linewidth]{\rule{\linewidth}{3.5pt}}

\begin{center}
	{\LARGE \bf Supplementary Information for:\\ \titl\par}
\end{center}
\noindent\makebox[\linewidth]{\rule{\linewidth}{1pt}}

\section{Architectures used for supervised painting tasks}
\seclabel{si_model_arch}

\figref{architecture} depicts architectures used in each of the two supervised tasks going from coordinates to images: Supervised Coordinate Classification (\secref{supervised_classification1}), and Supervised Rendering (\secref{supervised_rendering}).

In the case of convolution, or, in this case, transposed convolution (deconvolution), the same architecture is used for both tasks, as shown in the top row of \figref{architecture}, but we generally found the Supervised Rendering tasks requires wider layers (more channels). Top performing deconvolutional models in Supervised Coordinate Classification have $c=1$ or $2$, while in Supervised Rendering we usually need $c=2,3$. In terms of convolutional filter size, filter sizes of 2 and 4 seem to outperform 3 in Coordinate Classification, while in Rendering the difference is less distinctive.

Note that the CoordConv model only replaces the first layer with CoordConv (shown in green in \figref{architecture} ).






\figp[H]{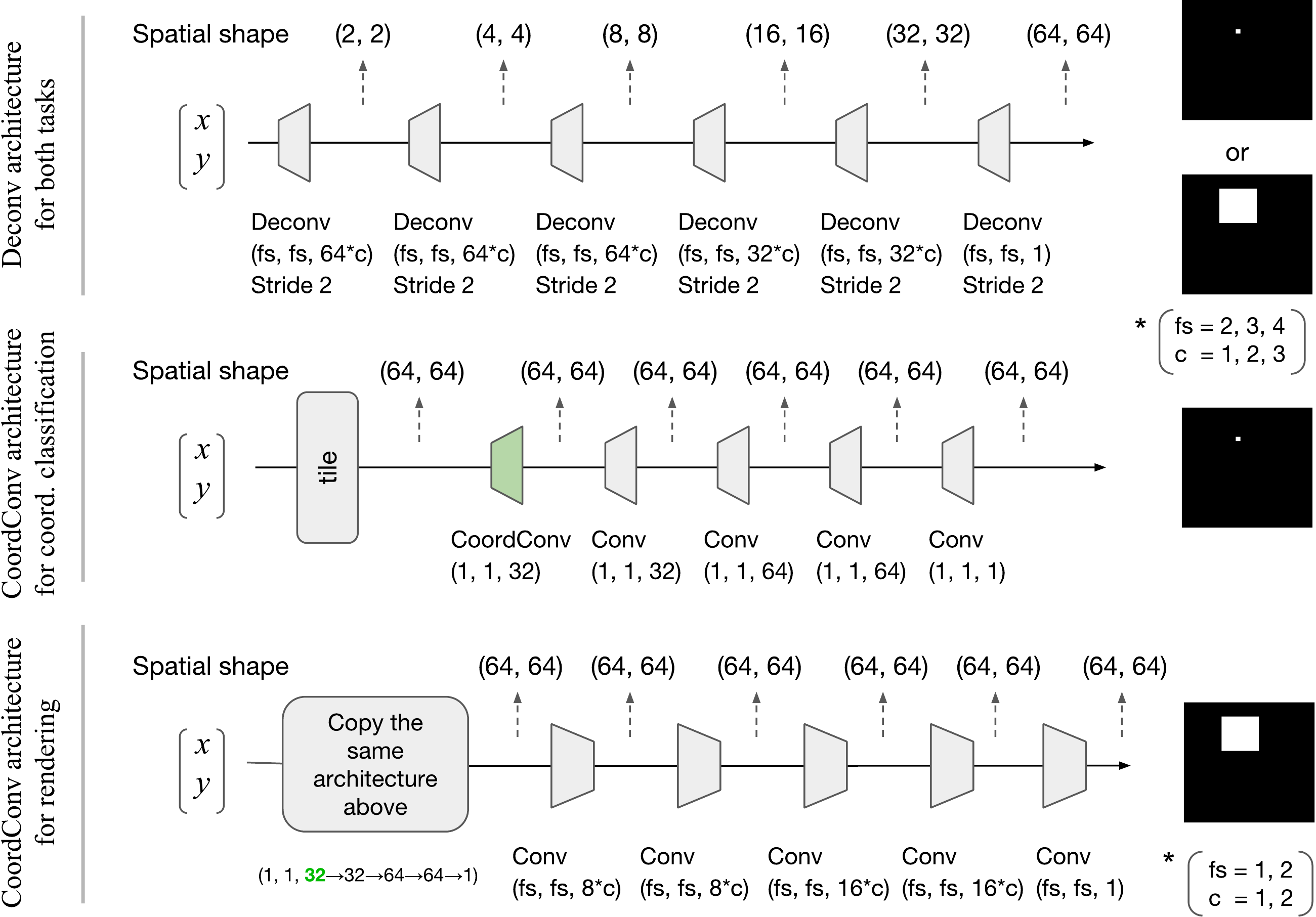}{.9}{Deconvolutional and CoordConv architectures used in each of the two supervised tasks. ``fs" stands for filter size, and ``c" for channel size. We use a grid search on different ranges of them as displayed underneath each model, while allowing deconvolutional models a wider range in both. Green indicates a CoordConv layer.}

\figp[H]{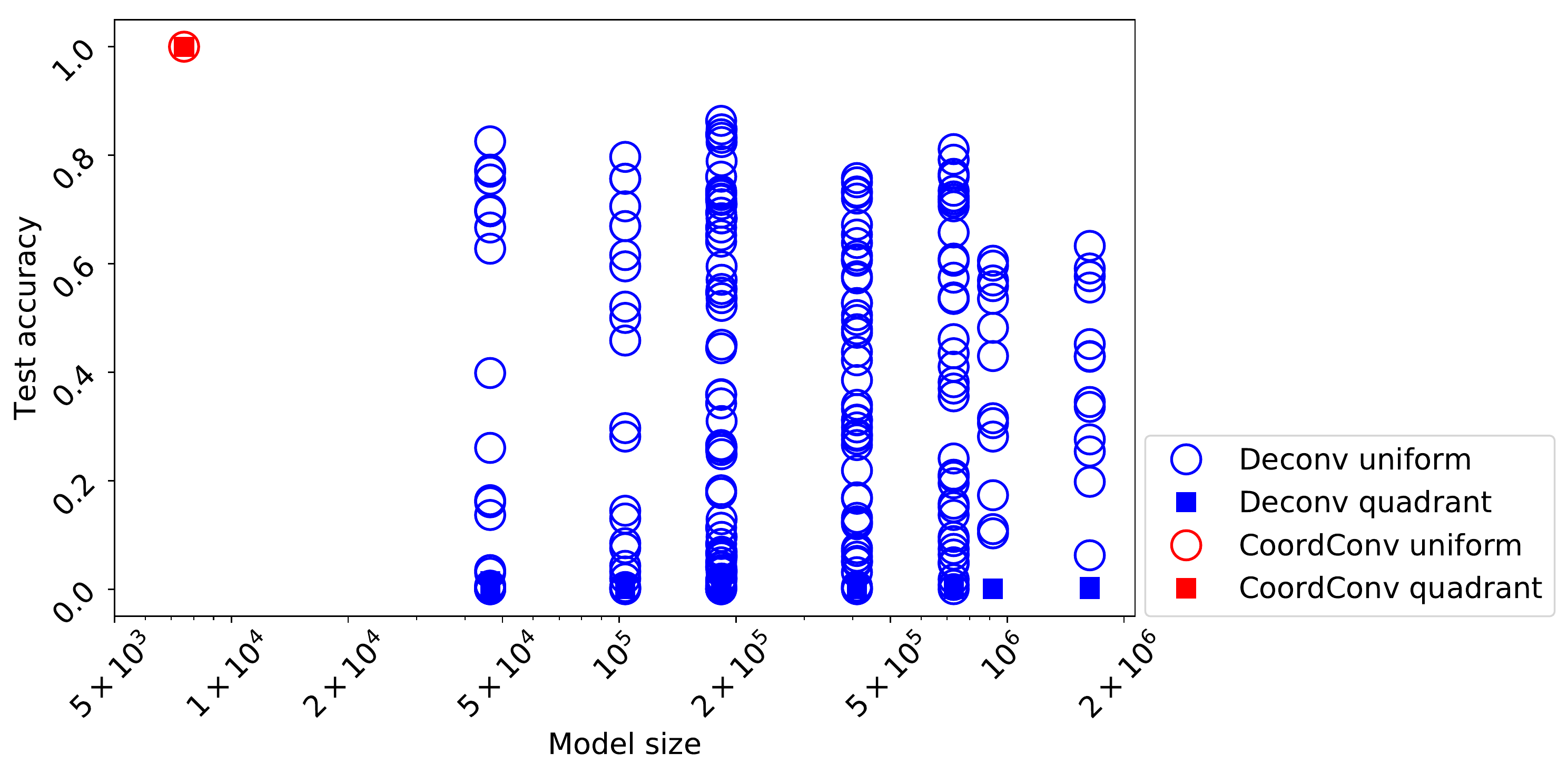}{.9}{Model size vs. test accuracy for the Supervised Coordinate Classification subtask on the uniform split and quadrant split. Deconv models (blue) of many sizes achieve 80\% or a little higher --- but never perfect --- test accuracy on the uniform split. On the quadrant split, while many models perform slightly better than chance (1/4096 = .000244) no model generalizes significantly. CoordConv model achieves perfect accuracy on both splits.}

Because of the usage of different filter sizes and channel sizes, we end up training models with a range of sizes. Each is combined with further grid searches on hyperparameters including the learning rate, weight decay, and minibatch sizes. 
Therefore at the same size we end up with multiple models with a spread of performances, as shown in \figref{supcoord_deconv_coordconv_modelsize} for the Supervised Coordinate Classification task. 
We repeat the same exact setting of experiments on both uniform and quadrant splits, which result in the same number of experiments. It is not obviously shown in \figref{supcoord_deconv_coordconv_modelsize} because quadrant trainings are mostly poorly (at the bottom of the figure).

As can be seen, it seems unlikely that even larger models would perform better. They all basically struggle to get to a good test accuracy. This (1) confirms that performance is not simply being limited by model size, as well as (2) shows that working CoordConv models are one to two orders of magnitude smaller (7553 as opposed to 50k-1.6M parameters) than the best convolutional models.
The model size vs. test performance plot on Supervised Rendering is similar (not shown), except CoordConv model in that case has a slightly larger number of parameters: 9490. CoordConv achieves perfect test IOU there while deconvolutional models struggle at sizes 183k to 1.6M.


\section{Further Supervised Coordinate Classification details}
\seclabel{si_supervised_coordinate}

For deconvolutional models, we use the model structure as depicted in the top row in \figref{architecture}, while varying the choice of filter size (\{2, 3, 4\}) and channel size multipliers (\{1,2,3\}), and each combined with a hyperparameter sweep of learning rate \{0.001, 0.002, 0.005, 0.01, 0.02, 0.05\}, and weight decay \{0.001, 0.01\}. Models are trained using a softmax output with cross entropy loss with Adam optimizer. We train 1000 epochs with minibatch size of 16 and 32. The learning rate is dropped to 10\% every 200 epochs for four times. \todo{Best model obtained is: xxxx}

For CoordConv models, because it converges so fast and easy, we did not have to try a lot of settings --- only 3 learning rates \{0.01 0.001, 0.005\} and they all learned perfectly well. There's also no need for learning rate schedules as it quickly converges in 10 seconds.

\figref{deconv_coordconv_uniform_2789} demonstrates how accurate and smooth the learned probability mass is with CoordConv, and not so much with Deconv. We first show the overall $64\times64$ map of logits, one for a training example and one for a test example just right next to the train. Then we zoom in to a smaller region to examine the intricacies. We can see that convolution, even though designed to act in a translation-invariant way, shows artifacts of not being able to accomplish so.

\figp[H]{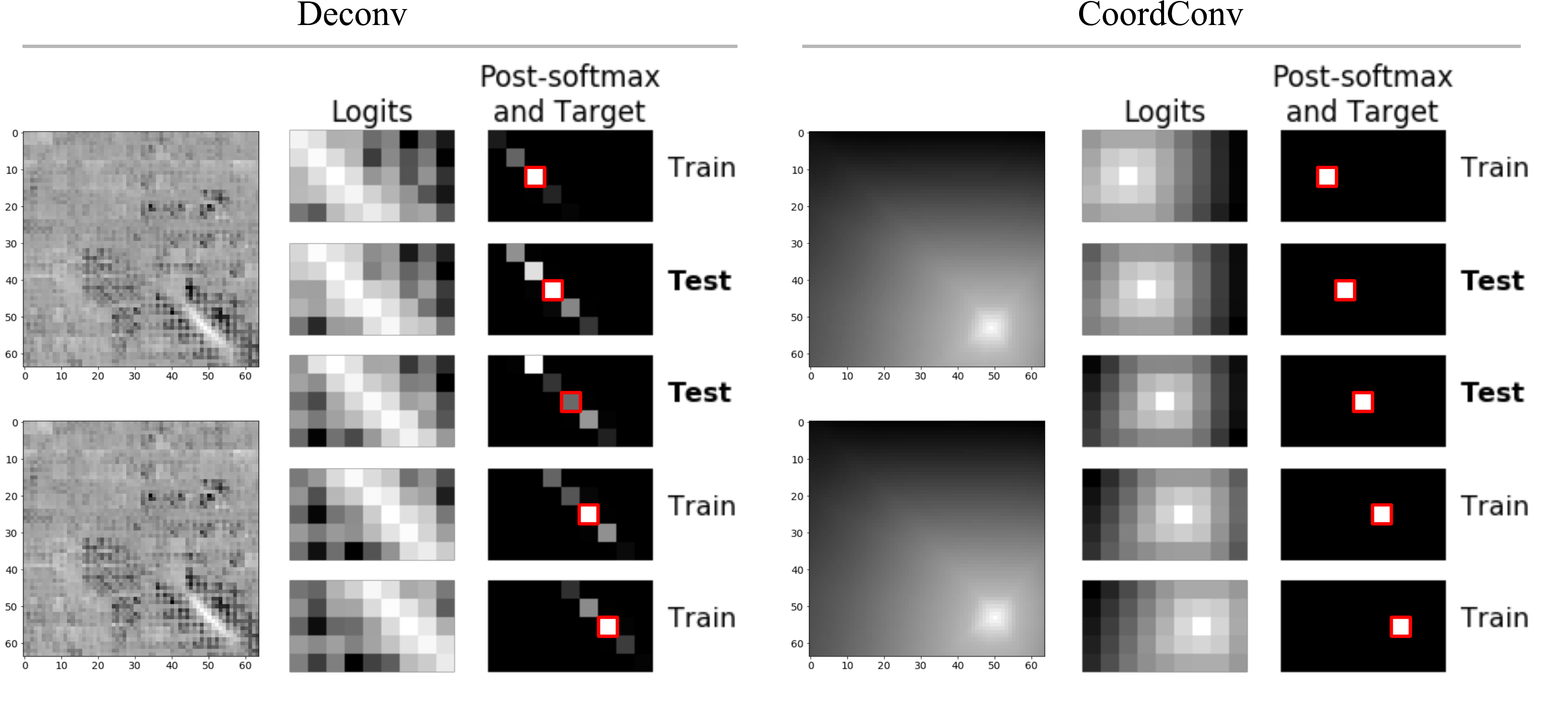}{1.}{Comparison of behaviors between Deconv model and CoordConv model on the Supervised Coordinate Classification task. We select five horizontally neighboring pixels, containing samples in both train and test splits, and zoom in on a $5\times9$ section of the $64\times64$ canvas so the detail of the logits and predicted probabilities may be seen. The full $64\times64$ map of logits of the first two samples (first in train, second in test) are also shown. The deconvolutional model outputs probabilities in a decidedly non-translation-invariant manner.}

\section{Further Supervised Coordinate Regression details}
\seclabel{si_supervised_regress}

Exact architectures applied in the Supervised Coordinate Regression task are described in table \ref{scr_table}.
For the uniform split, the best-generalizing convolution architecture consisted of a stack of alternating
convolution and max-pooling layers, followed by a fully-connected layer and an output layer. This architecture
was fairly robust to changes in hyperparameters. In contrast, for the quandrant split, the best-generalizing
network consisted of strided convolutions feeding into a global-pooling output layer, and good performance
was delicate. In particular, training and generalization was sensitive to the number of batch normalization layers (2), weight decay strength (5e-4), and optimizer (Adam, learning rate 5e-4). A single CoordConv architecture generalized perfectly with the same hyperparameters over both splits, and consisted of a single CoordConv layer followed by additional layers of convolution, feeding into a global pooling output layer.


\begin{table}[h]
\caption{Model Architectures for Supervised Coordinate Regression. FC: Fully-connected, MP: Max Pooling, GP: Global Pooling, BN: Batch normalization, s2: stride 2.}
\centering
\begin{tabu} { p{1.5cm}| p{6.5cm} | p{4.5cm} }
\toprule
	&  Conv & CoordConv \\
\hline
\hline
	Uniform Split & 3$\times$3, 16 - MP 2$\times$2 - 3$\times$3, 16 - MP 2$\times$2 - 3$\times$3, 16 - MP 2$\times$2 - 3$\times$3, 16 - FC 64 - FC 2 &  \multirow{2}{4.5cm}{1$\times$1, 8 - 1$\times$1, 8 - 1$\times$1, 8 - 3$\times$3, 8 - 3$\times$3, 2 - GP} \\
\cline{1-2}
	Quadrant Split & 5$\times$5 (s2), 16 - 1$\times$1, 16 - BN - 3$\times$3, 16 - 3$\times$3 (s2), 16 - 3$\times$3 (s2), 16 - BN - 3$\times$3 (s2), 16 - 1$\times$1, 16 - 3$\times$3 (s2), 16 - 3$\times$3, 2 - GP  & \\
\hline
\end{tabu}
\label{scr_table}
\end{table}

\section{Further Supervised Rendering details}
\seclabel{si_supervised_rendering}

Both the architectural and experimental settings are similar to \secref{si_supervised_coordinate} except the loss used is pixelwise sigmoid output with cross entropy loss. We also tried mean squared error loss but the performance is even weaker. We performed heavy hyperparameter sweeping and deliberate learning rate annealing for Deconv models (same as said in \secref{si_supervised_coordinate}), while in CoordConv models it is fairly easy to find a good setting. All models trained with learning rates \{0.001, 0.005\}, weight decay \{0, 0.001\}, filter size \{1, 2\} turned out to perform well after 1--2 minutes of training. Take the best model obtained,
\figref{suppaint_heatmap_uniform} and \figref{suppaint_heatmap_quadrant} show the learned logits and pixelwise probability distributions for three samples each, in the uniform and quadrant cases, respectively. 
We can see that the CoordConv model learns a much smoother and precise distribution. All samples are test samples.

\figp[H]{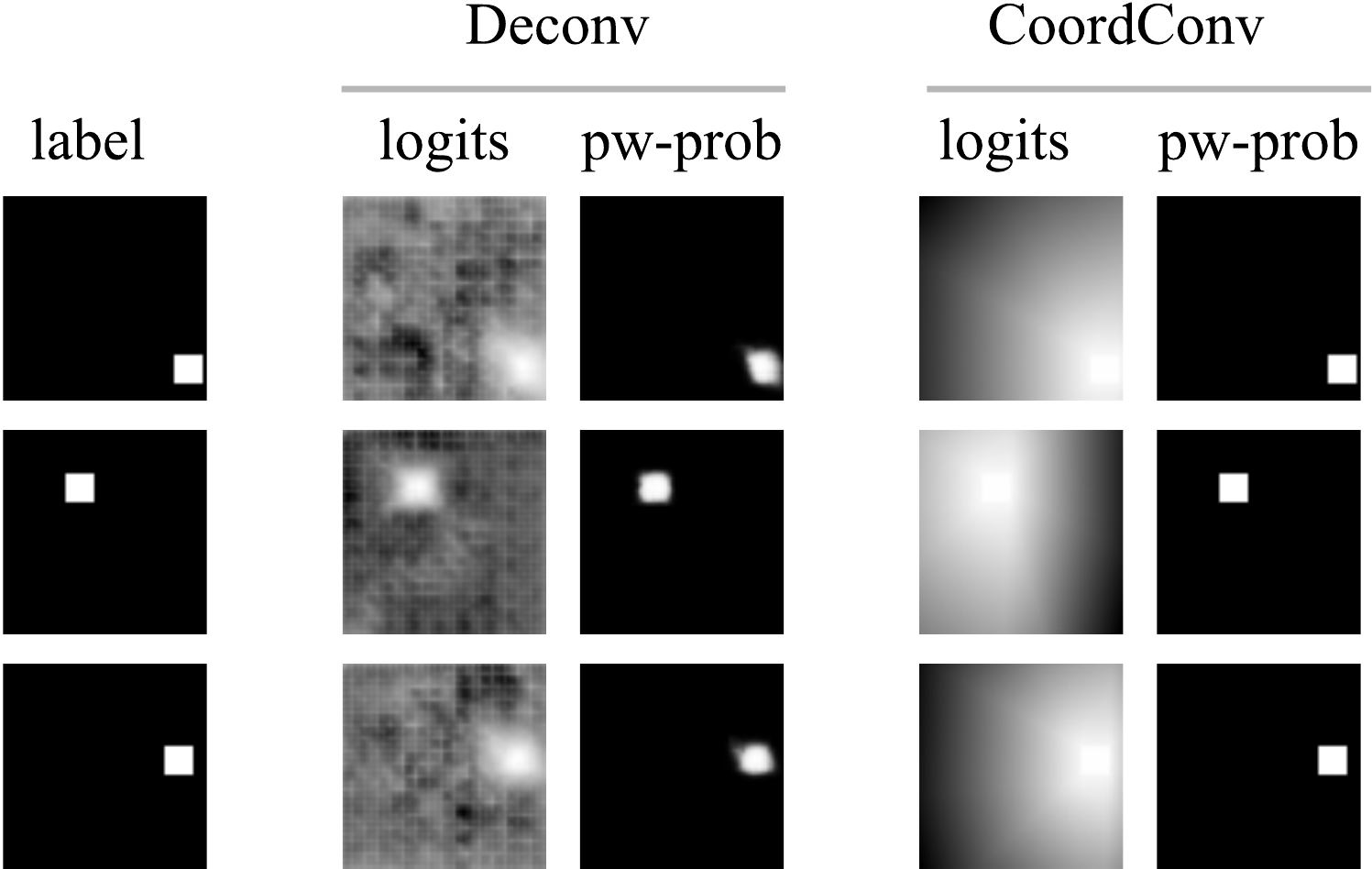}{.6}{Output comparison between Deconv and CoordConv models on three test samples. Models are trained on a \emph{uniform} split. Logits are model's direct output; pixelwise probability (pw-prob) is logits after Sigmoid. Conv outputs (middle columns) manage to get roughly right. CoordConv outputs (right columns) are precisely correct and its logit maps are smooth.\todo{Best conv model: xxxx; best CoordConv model:}}

\figp[H]{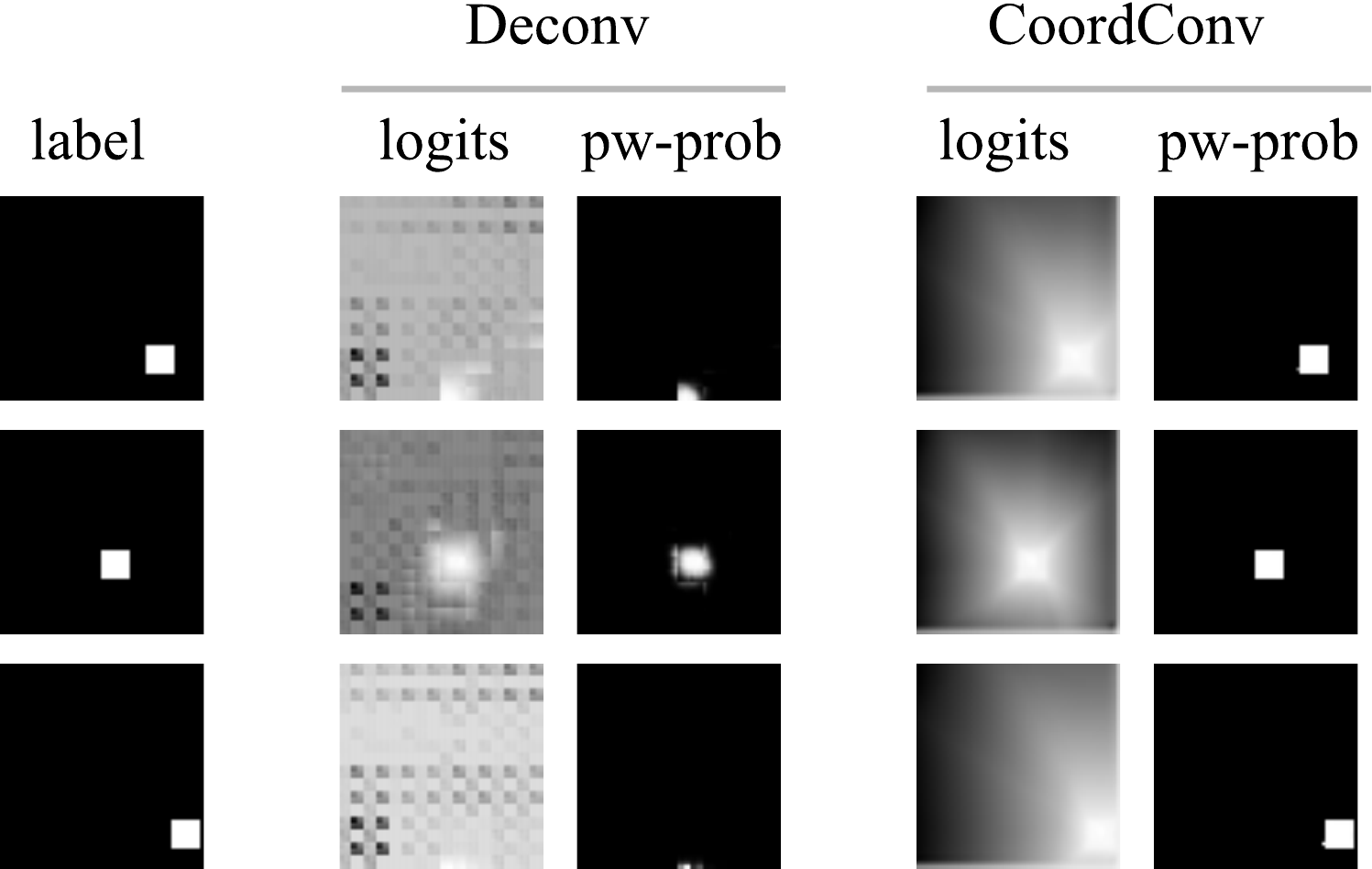}{.6}{Output comparison between Deconv and CoordConv models on three test samples. Models are trained on a \emph{quadrant} split. Logits are model's direct output; pixelwise probability (pw-prob) is logits after Sigmoid. Conv outputs (middle columns) failed mostly. Even with such a difficult generalization problem, CoordConv outputs (right columns) are precisely correct and its logit maps are smooth.\todo{Best conv model: xxxx; best CoordConv model:}}

\section{Further ImageNet classification details}
\seclabel{si_imagenet}

We evaluate the potential of CoordConv in image classification with ImageNet experiments. We take ResNet-50 and run the baseline on distributed framework using 100 GPUs, with the open-source framework Horovod. For CoordConv variants, we add an extra CoordConv layer only in the beginning, which takes a 6-channel tensor containing image RBG, $i$, $j$ coordinates and pixel distance to center $r$, and output 8 channels with 1$\times$1 convolution. The increase of parameters is negligible. It then goes in with the rest of ResNet-50.

Each model is run 5 times on the same setting to account for experimental variances. Table.~\ref{imagenet_numbers} lists the test result from each run in the end of 90 epochs. CoordConv model obtains better average result on two of the three measures, however a one-sided t-test tells that the improvement on Top 5 accuracy is not quite statistically significant with $p=.11$. 

Of all vision tasks, we might expect image classification to show the least performance change when using CoordConv instead of convolution, as classification is more about what is in the image than where it is. This tiny amount of improvement validates that.

\begin{table}[h]
\caption{ImageNet classification result comparison between a baseline ResNet-50 and CoordConv ResNet-50. For each model three experiments are run, listed in three separate rows below.
}
\centering
\begin{tabu} { c| c | c | c}
\toprule
& Test loss & Top-1 Accuracy & Top-5 Accuracy \\
\hline
\hline
\multirow{3}{2cm}{Baseline ResNet-50} & 1.43005 & 0.75722 & 0.92622\\
& 1.42385 & 0.75844 & 0.9272\\
& 1.42634 & 0.75782 & 0.92754\\
& 1.42166 & 0.75692	& 0.92756\\
& 1.42671 & 0.75724	& 0.92708\\
\hline
Average & 1.425722 & \textbf{0.757528} & 0.92712 \\
\hline \hline

\multirow{3}{2cm}{CoordConv ResNet-50} & 1.42335 & 0.75732 & 0.92802\\
& 1.42492 & 0.75836 & 0.92754\\
& 1.42478 & 0.75774 & 0.92818\\
& 1.42882 & 0.75702 & 0.92694\\
& 1.42438 & 0.75668 & 0.92714\\
\hline
Average & \textbf{1.42525} & 0.757424 & \textbf{0.927564} \\
\hline \hline

\end{tabu}
\label{imagenet_numbers}
\end{table}

\section{Further object detection details}
\seclabel{si_objdet}

The object detection experiments are on a dataset containing randomly rescaled and placed MNIST digits on a $64\times64$ canvas. 
To make it more akin to natural images, we generate a much larger canvas and then center crop it to be $64\times64$, so that digits can be partially outside of the canvas.
We kept images that contain 5 digit objects whose centers are within the canvas. In the end we use 9000 images as the training set and 1000 as test.

A schematic of the model architecture is illustrated in \figref{rpn}. We use number of anchors $A=9$, with sizes $(15, 15)$, $(20, 20)$, $(25, 25)$, $(15, 20)$, $(20, 25)$, $(20, 15)$, $(25, 20)$, $(15, 25)$, $(25, 15)$. In box sampling (training mode), $p\_size$ and $n\_size$ are 6. In box non-maximum suppression (NMS) (test mode), the IOU threshold is 0.8 and maximum number of proposal boxes is 10. After the boxes are proposed and shifted, we do not have a downstream classification task, but just calculate the loss from the boxes. The training loss include box loss and score loss. As evaluation metric we also calculate IOUs between proposed boxes and ground truth boxes. Table.~\ref{rpn_numbers} lists those metrics obtained the test dataset, by both Conv and CoordConv models. We found that every metric is improved by CoordConv, and the average test IOU improved by about 24 percent.

\figp[H]{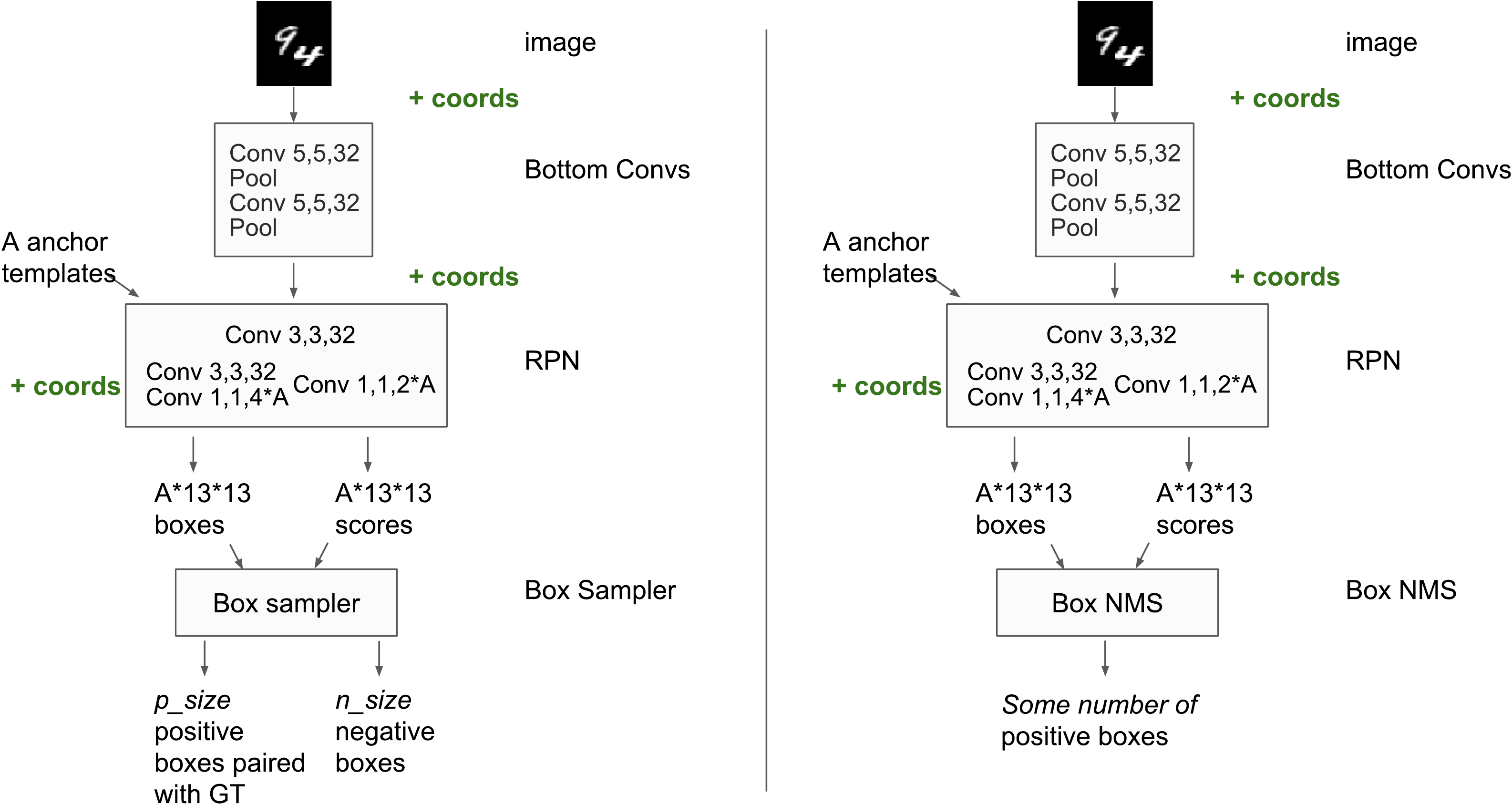}{.8}{Faster R-CNN architecture used for object detection on scattered MNIST digits. Green indicates where coordinates are added. Note that the input image is used for demonstration purpose. The real dataset contains 5 digits on a canvas and allows overlapping.
 \textbf{(Left)} train mode with box sampler. \textbf{(Right)} test mode with box NMS.}


\begin{table}[h]
\caption{MNIST digits detection result comparison between a Faster R-CNN model with regular convolution vs. with CoordConv. Metrics are all on test set. Train IOU: average IOU between sampled positive boxes (train mode) and ground truth; Test IOU-average): average IOU between 10 selected boxes (test mode) and ground truth; Test IOU-select: average IOU between the best scored box and its closest ground truth.}

\centering
\begin{tabu} { l| c | c | c}
\toprule
& Conv & CoordConv & \% Improvement \\
\hline
\hline
Box loss & 0.1003 & 0.0854 & 17.44\\
Score loss & 0.5270 &  0.2526  & 108.63\\
Total loss (sum of the two above) & 0.6271 & 0.3383 & 85.37\\
Train IOU & 0.6388 &  0.6612  & 3.38\\
Test IOU-average & 0.1508 &  0.1868  & 23.87\\
Test IOU-select &  0.4965 &  0.6359  & 28.08\\
\hline \hline

\end{tabu}
\label{rpn_numbers}
\end{table}

\section{Further generative modeling details}

\subsection{GANs on colored shapes}
\seclabel{si_gan_clevr}

\paragraph{Data.} The dataset used to train the generative models is 50k red-and-blue-object images of size $64\times64$. We follow the same mechanism as Sort-of-Clevr, in that objects appear at random positions on a white background without overlapping, only limiting the number of objects to be 2 per image. The objects are always one red and one blue, of a randomly chosen shape out of \{circle, square\}. Examples of images from this dataset can be seen in the top row, leftmost column in \figref{gan_samples}, at the intersection of ``Real images" and  ``Random samples".

\paragraph{Architecture and training details.} The $z$ dimension to both regular GAN and CoordConv GAN is 256. In GAN, the generator uses 4 layers of deconvolution with strides of 2 to project $z$ to a $64\times64$ image shape. The parameter size of the generator is 6,413,315. In CoordConv GAN, we add coordinate channels only at the beginning, making the first layer CoordConv, and then continue with normal Conv.. The generator in this case uses mostly (1,1) convolutions and has only 444,931 parameters. The same discriminator is used for both models. In the case where we also turn the discriminator to be CoordConv like, its first Conv layer is replaced by CoordConv, and the parameter size increases from 4,316,545 to 4,319,745. The details of both architectures can be seen in Table.~\ref{gan_archs}. We trained two CoordConv GAN versions where CoordConv applies: 1) only in generator, and 2) in both generator and discriminator. They end up performing similarly well. The demonstrated examples in all figures are from one in the latter case.

The change needed to make a generator whose first layer is fully-connected CoordConv is trivial. Instead of taking image tensors which already have Cartesian dimensions, the CoordConv generator first tiles $z$ vector into a full $64\times64$ space, and then concatenate it with coordinates in that space.

To train each model we use a fixed learning rate 0.0001 for the discriminator and 0.0005 for the generator. In each iteration discriminator is trained once followed by generator trained twice. The random noise vector $z$ is drawn from a uniform distribution between $[-1,1]$. We train each model for 50 epochs and save the model in the end of every epoch. We repeat the training with the same hyperparameters 5 to 10 times for each, and pick the best model for each to show a fair comparison in all figures.





\begin{table}[h]
\caption{Model Architectures for GAN and CoordConv GAN used in the colored shape generation. In the case of CoordConv GAN, only the first layer is changed from regular Conv to CoordConv. FC: fully connected layer; s2: stride 2. 
}
\centering
\begin{tabu} { p{1.5cm}| p{6.5cm} | p{4cm} }
\toprule
& Generator & Discriminator \\
\hline
\hline
GAN & FC 8192 (reshape 4$\times$4$\times$512) - 5$\times$5, 256 (s2) - 5$\times$5, 128 (s2) - 5$\times$5, 64 (s2) - 5$\times$5, 3 (s2) - Tanh & \multirow{2}{4cm}{5$\times$5, 64 (s2) - 5$\times$5, 128 (s2) - 5$\times$5, 256 (s2) - 5$\times$5, 512 (s2) - 1} \\

\cline{1-2}
CoordConv GAN & 1$\times$1, 512 - 1$\times$1, 256 - 1$\times$1, 256 - 1$\times$1, 128 - 1$\times$1, 64 - 3$\times$3, 64 - 3$\times$3, 64 - 1$\times$1, 3 & \\

\hline
\end{tabu}
\label{gan_archs}
\end{table}

\paragraph{Latent interpolation.} Latent interpolation is conducted by randomly choosing two noise vectors, each from a uniform distribution, and linearly interpolate in between with an $\alpha$ factor that indicates how close it is to the first vector. \figref{gan_interpolate} and \figref{coordgan_interpolate} each show, on regular GAN and CoordConv GAN, respectively, five random samples of pairs to conduct interpolation with. In addition to \figref{coordgan_interpolate},  \figref{coordgan_interpolate_special} shows deliberately picked examples that exhibit a special moving effect that has only been seen in CoordConv GAN.

\figp[h]{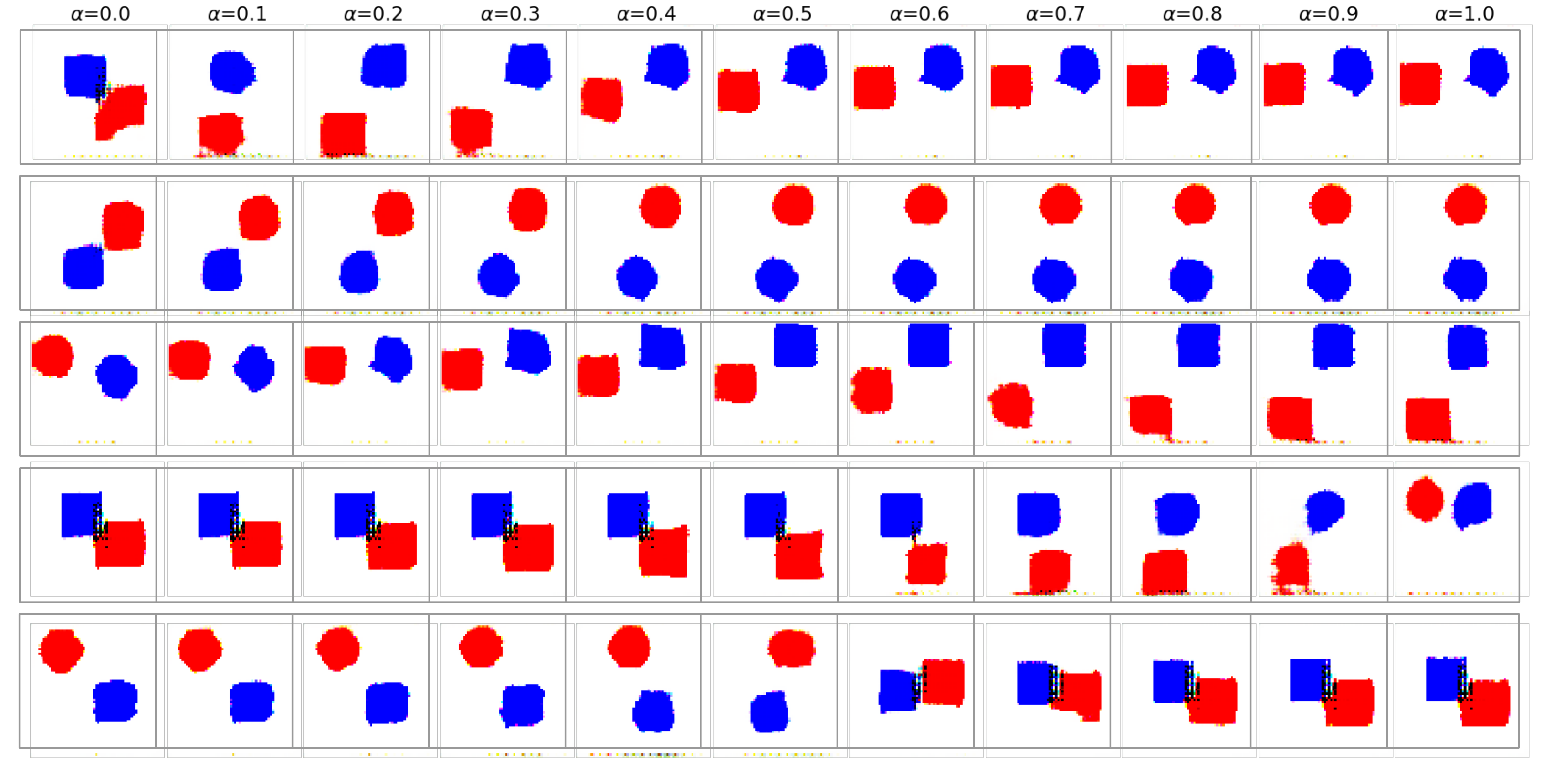}{.9}{Regular GAN samples with a series of interpolation between 5 random pairs of $z$. Also observed position and shape transitioning but are different.}

\figp[h]{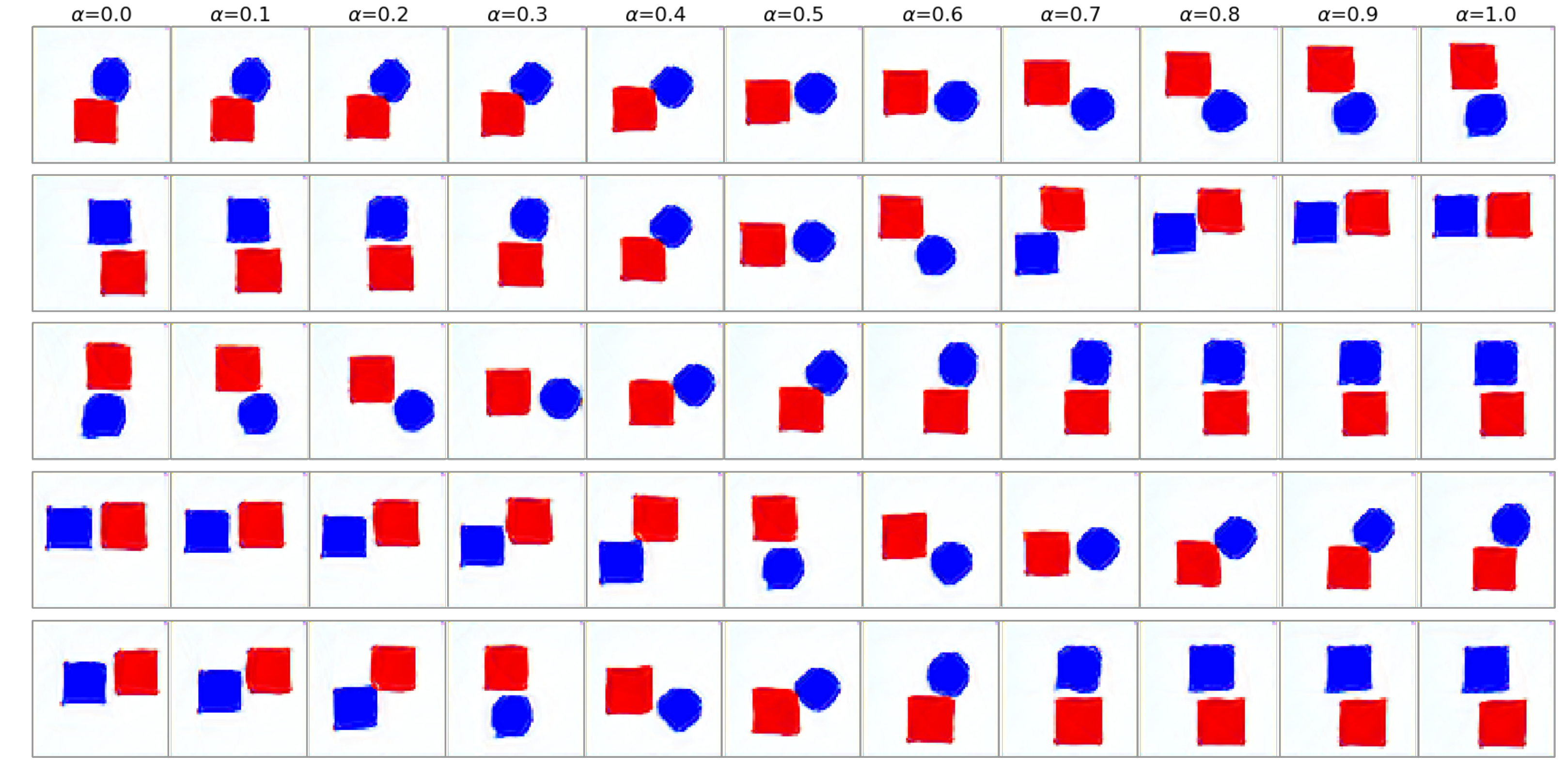}{.9}{CoordConv GAN samples with a series of interpolation between 5 random pairs of $z$. Top row: at the position in the manifold, the model has learned a smooth circular motion. The rest of the rows: the circular relation between two objets is still observed, while some object shapes also undergo a smooth transition.
}

\figp[h]{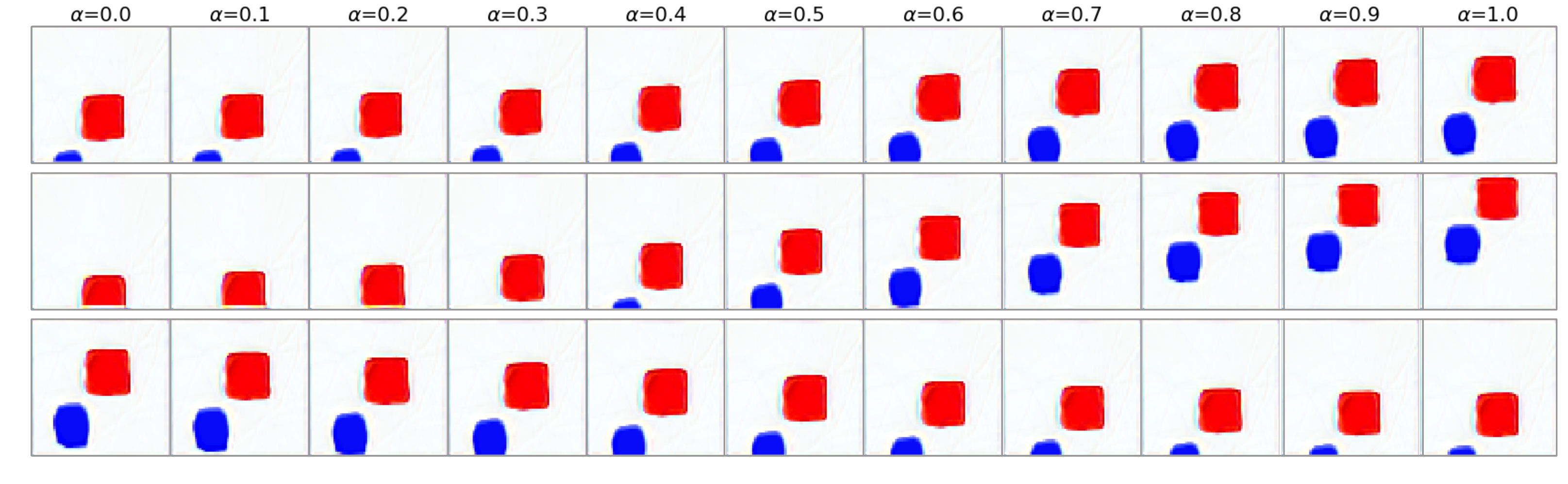}{.9}{A special effect only observed in CoordConv GAN latent space interpolation: two objects stay constant in relative positions to each other but move together in space. They even move out of the scene which is never present in the real data ---  learned to extrapolate. These 3 examples are picked from many random drawings of $z$ pairs, as opposed to \figref{coordgan_interpolate} and \figref{gan_interpolate}, where first 5 random drawings are shown.}

\figp[h]{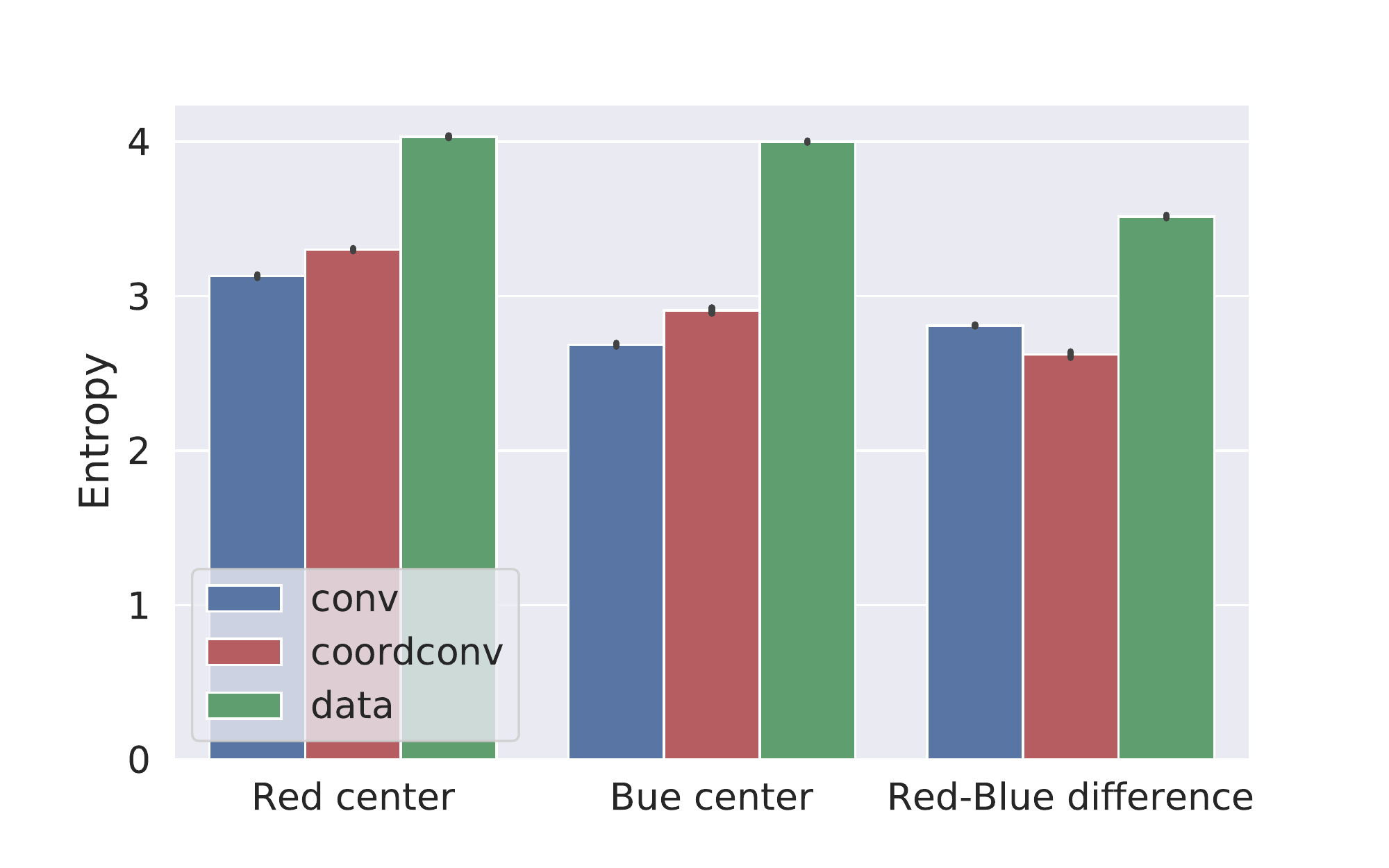}{.7}{Entropy values and confidence intervals of the sampled results in \figref{gan_samples}, column (b) and (c).}

\paragraph{Measure of entropy.}  
In \figref{gan_samples}, we reduce generated red and blue objects to their centers and plot the coverage of space in column (b) and relative locations in (c). To make the comparison quantitative, we can further calculate the entropy in each case, reducing each figure in (b) and (c) to an entropy value shown as a bar in \figref{entropy_bar}. Confidence intervals of each bar is also shown by repeating the experiment 10 times. We can see that CoordConv (red) is closer to data (green) in objects' coverage of space, but has more of a mode collapse in objects' relative position.

\subsection{VAEs on colored shapes}
\seclabel{si_vae}

We train both convolutional and CoordConv VAEs on the same dataset of 50k 64 x 64 images of blue and red non-overlapping squares and circles, as described in \secref{si_gan_clevr}.
Convolutional VAEs exhibit many of the same problems that we observed in GANs, and adding CoordConv confers many of the same benefits.


A VAE is composed of an encoder that maps data to a latent and a decoder that maps the latent back to data. With minor exceptions our VAE's encoder architecture is the same as our GAN's discriminator and it's decoder is the same as our GAN's generator. The important difference is of course that the output shape of the encoder is the size of the latent (in this case 8), not two as in a discriminator.

Architectures are shown in Table.~\ref{vae_archs}. The decoder architectures of the convolutional control and CoordConv experiments are similar aside from kernel size - the CoordConv decoder uses 1x1 kernels while the convolutional decoder uses 5x5 kernels. 

Due to the pixel sparsity of images in the dataset we found it important to weight reconstruction loss more heavily than latent loss by a factor of 50. Doing so didn't interfere with the quality of the encoding. We used Adam with a learning rate of 0.005 and no weight decay.

\begin{table}[ht]
	\caption{Model Architectures: Convolutional VAE and CoordConv VAE
	}
	\centering
	\begin{tabu} { p{1.5cm}| p{6.5cm} | p{4cm} }
		\toprule
		& Decoder & Encoder \\
		\hline
		\hline
		VAE & FC 8192 (reshape 4$\times$4$\times$512) - 5$\times$5, 256 (s2) - 5$\times$5, 128 (s2) - 5$\times$5, 64 (s2) - 5$\times$5, 3 (s2) - Sigmoid & \multirow{2}{4cm}{5$\times$5, 64 (s2) - 5$\times$5, 128 (s2) - 5$\times$5, 256 (s2) - 5$\times$5, 512 (s2) - Flatten - FC, 10} \\
		
		\cline{1-2}
		CoordConv VAE & 1$\times$1, 512 - 1$\times$1, 256 - 1$\times$1, 256 - 1$\times$1, 128 - 1$\times$1, 64 - 1$\times$1, 3 - Sigmoid\\
		
		\hline
	\end{tabu}
	\label{vae_archs}
\end{table}

\figp[H]{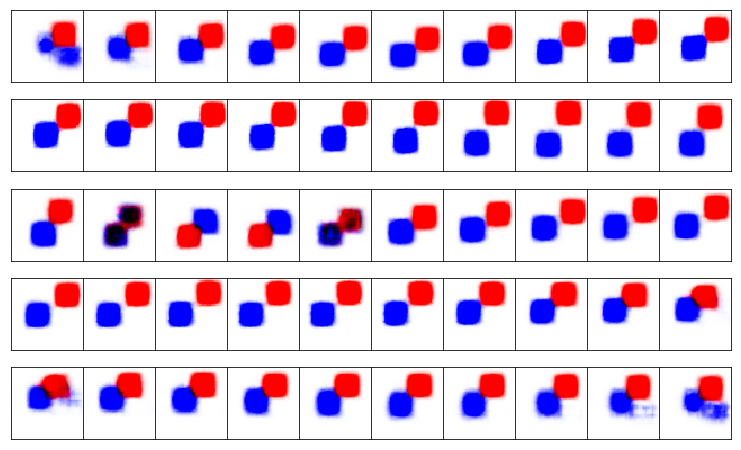}{.9}{Latent space interpolations from a VAE without CoordConv. The red and blue shapes are mostly stationary. When they do move they do so by disappearing and appearing elsewhere in pixel space. Smooth changes in the latent don't translate to smooth geometric changes in pixel space. The latents we interpolated between were sampled randomly from a uniform distribution.}

\figp[H]{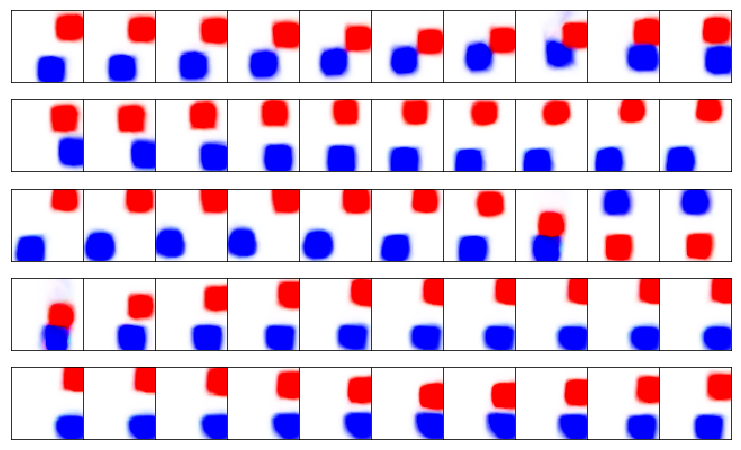}{.9}{Latent space interpolations from a VAE with CoordConv in the encoder and decoder. The red and blue shapes span pixel space more fully and smooth changes in latent space map to smooth changes in pixel space. Like the CoordConv GAN, the CoordConv VAE is able to extrapolate beyond the borders of the frame it was trained on. The latents we interpolated between were sampled randomly from a uniform distribution.}

\subsection{GANs on LSUN}
\seclabel{si_gan_lsun}
The dataset used to train the generative models is LSUN bedroom, composed of 3,033,042 images of size $64\times64$.

The architectures adopted (see Table.~\ref{lsun_gan_archs}) are similar to the ones adopted for generating the colored shape results in  \secref{si_gan_clevr}, with a few noticeable differences:

\begin{itemize}
\item We use CoordConv layers instead of regular Conv layers not only in the first layer of the discriminator, but in each layer.
\ $z$ is of dimension 100.
\item The GAN generator includes a layer mapping from $z$ to a 4x4x1024 tensor and the other layers have double the number of channels.
\item CoordConv GAN generator has more channels for each layer.
\end{itemize}

\begin{table}[h]
	\caption{Model Architectures for GAN and CoordConv GAN for LSUN. FC: fully connected layer; s2: stride 2. 
	}
	\centering
	\begin{tabu} { p{1.5cm}| p{6.5cm} | p{4cm} }
		\toprule
		& Generator & Discriminator \\
		\hline
		\hline
		GAN & FC 16384 (reshape 4$\times$4$\times$1024) - 5$\times$5, 512 (s2) - 5$\times$5, 256 (s2) - 5$\times$5, 128 (s2) - 5$\times$5, 3 (s2) - Tanh & \multirow{2}{4cm}{5$\times$5, 64 (s2) - 5$\times$5, 128 (s2) - 5$\times$5, 256 (s2) - 5$\times$5, 512 (s2) - 1} \\
		
		\cline{1-2}
		CoordConv GAN & 1$\times$1, 1024 - 1$\times$1, 512 - 1$\times$1, 256 - 1$\times$1, 256 - 1$\times$1, 128 - 3$\times$3, 128 - 3$\times$3, 64 - 1$\times$1, 3 & \\
		
		\hline
	\end{tabu}
	\label{lsun_gan_archs}
\end{table}

\figp[H]{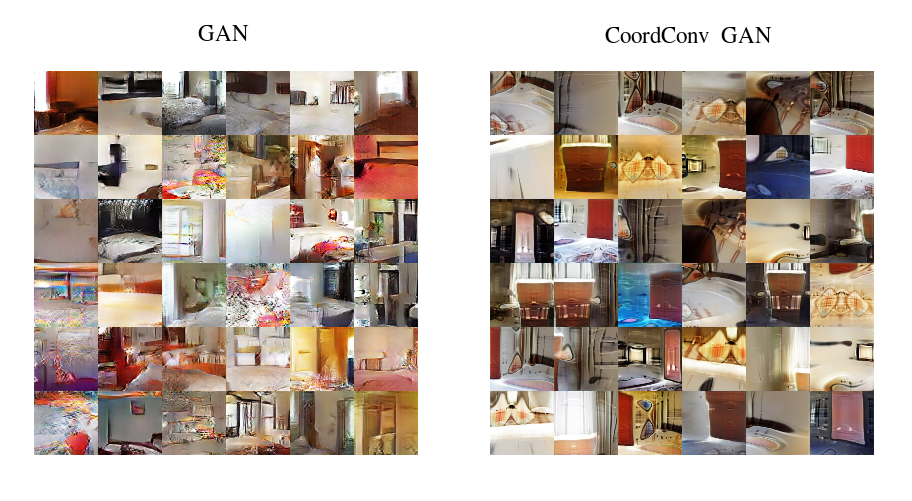}{.9}{Samples from the regular GAN (left) and the CoordConv GAN (right).}

Samples from both models are provided in \figref{gans_lsun_samples}.
One peculiar property of the CoordConv GAN model with respect to the regular GAN one is the geometric interpolation.
As shown in Figure~\figref{gan_lsun_interpolate} in regular GAN interpolations objects appear and disappear, while in CoordConv GAN interpolations in \figref{coordconv_gan_lsun_interpolate} objects move around, translating, enlarging, squashing and doing geometric transformations over them.

\figp[H]{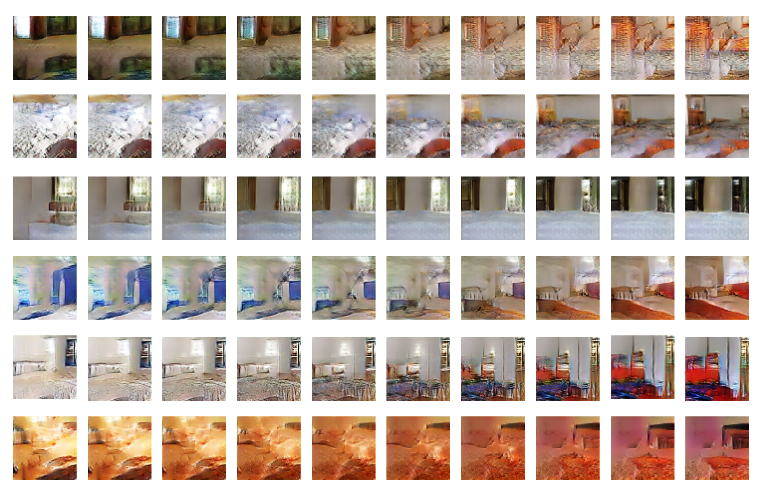}{.9}{Samples of regular GAN trained on LSUN with a series of interpolation between 5 random pairs of $z$.}

\figp[H]{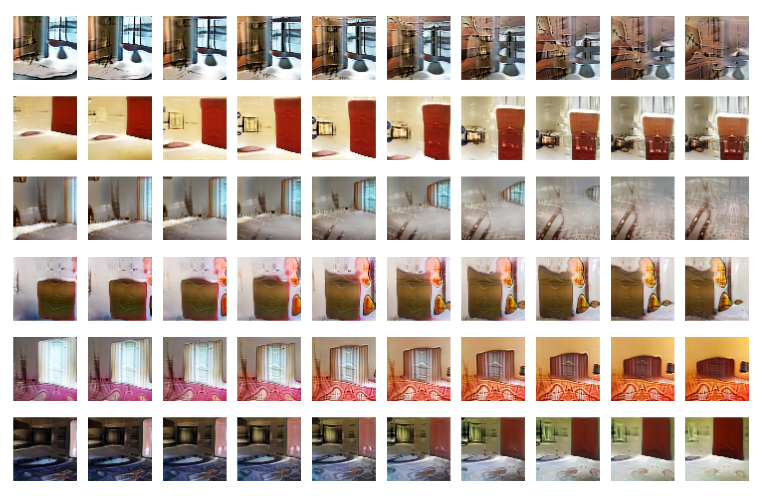}{.9}{Samples of CoordConv GAN trained on LSUN with a series of interpolation between 5 random pairs of $z$.}

The regular GAN has been trained for 11000 steps of batch size 128, while the CoordConv GAN has been trained 22000 steps of batch size 64 (because the available memory on the GPUs did not allow for 128).
Both models have been trained using Horovod to distribute the training on 16 GPUs.

\section{Further reinforcement learning details}
\seclabel{si_rl}

We used OpenAI baselines \footnote{https://github.com/openai/baselines/} implementation and default parameters on all experiments. Table.~\ref{game_scores} shows the average scores obtained at the end of game over 10 runs of each.

\begin{table}[h]
\caption {All games with final scores and p-values.}
\centering
\begin{tabu}{ c | c | c | c }
\hline 
Game & Conv & CoordConv & p-value \\
\hline \hline
Alien & 1462.5 & 2005.0 & 0.0821 \\ 
Bank Heist & 932.5 & 1330.0 & 0.1736 \\  
Ms. Pacman & 2557.5 & 3945.0 & 0.0065 \\
Robotank & 2.75 & 3.5 & 0.2899 \\
Centipede & 3359.5 & 3424.5 & 0.8703\\   
Asterix & 16250.0 & 35300.0 & 0.0003 \\
Asteroids & 2082.5 &1912.5 & 0.1124 \\
Amidar &1092.75 & 1137.5 & 0.2265\\
Seaquest & 1780.0 & 1780.0 & 0.4057\\
\hline
\end{tabu}
\label{game_scores}
\end{table}

\section{The CoordConv layer implementation}
\seclabel{si_layer}

\definecolor{verbgray}{gray}{0.9}

\lstdefinestyle{graystyle}{
	backgroundcolor=\color{verbgray},
	frame=single,
  	framerule=0pt,
 	basicstyle=\ttfamily,
  	columns=fullflexible}
	
\lstset{style=graystyle}

\begin{lstlisting}[language=Python]
from tensorflow.python.layers import base
import tensorflow as tf

class AddCoords(base.Layer):
    """Add coords to a tensor"""
    def __init__(self, x_dim=64, y_dim=64, with_r=False):
        super(AddCoords, self).__init__()
        self.x_dim = x_dim
        self.y_dim = y_dim
        self.with_r = with_r

    def call(self, input_tensor):
        """
        input_tensor: (batch, x_dim, y_dim, c)
        """
        batch_size_tensor = tf.shape(input_tensor)[0]  
        xx_ones = tf.ones([batch_size_tensor, self.x_dim], 
                          dtype=tf.int32)                       
        xx_ones = tf.expand_dims(xx_ones, -1)                  
        xx_range = tf.tile(tf.expand_dims(tf.range(self.y_dim), 0), 
                            [batch_size_tensor, 1])           
        xx_range = tf.expand_dims(xx_range, 1)               

        xx_channel = tf.matmul(xx_ones, xx_range)
        xx_channel = tf.expand_dims(xx_channel, -1) 

        yy_ones = tf.ones([batch_size_tensor, self.y_dim], 
                          dtype=tf.int32) 
        yy_ones = tf.expand_dims(yy_ones, 1)
        yy_range = tf.tile(tf.expand_dims(tf.range(self.x_dim), 0),
                            [batch_size_tensor, 1])
        yy_range = tf.expand_dims(yy_range, -1)

        yy_channel = tf.matmul(yy_range, yy_ones)
        yy_channel = tf.expand_dims(yy_channel, -1)

        xx_channel = tf.cast(xx_channel, 'float32') / (self.x_dim - 1)
        yy_channel = tf.cast(yy_channel, 'float32') / (self.y_dim - 1)
        xx_channel = xx_channel*2 - 1
        yy_channel = yy_channel*2 - 1

        ret = tf.concat([input_tensor, 
                         xx_channel, 
                         yy_channel], axis=-1)

        if self.with_r:
            rr = tf.sqrt( tf.square(xx_channel)
                    + tf.square(yy_channel)
                    )
            ret = tf.concat([ret, rr], axis=-1)

        return ret

class CoordConv(base.Layer):
    """CoordConv layer as in the paper."""
    def __init__(self, x_dim, y_dim, with_r, *args,  **kwargs):
        super(CoordConv, self).__init__()
        self.addcoords = AddCoords(x_dim=x_dim, 
                                   y_dim=y_dim, 
                                   with_r=with_r)
        self.conv = tf.layers.Conv2D(*args, **kwargs)

    def call(self, input_tensor):
        ret = self.addcoords(input_tensor)
        ret = self.conv(ret)
        return ret
\end{lstlisting}


\end{document}